%% file: main.tex
\documentclass[lettersize,journal]{IEEEtran}
\usepackage{amsmath,amsfonts}
\usepackage{algorithmic}
\usepackage{array}
\usepackage[caption=false,font=normalsize,labelfont=sf,textfont=sf]{subfig}
\usepackage{textcomp}
\usepackage{stfloats}
\usepackage{url}
\usepackage{verbatim}
\usepackage{graphicx}
\usepackage{amssymb}
\usepackage{multirow}
\usepackage{makecell}

\usepackage[switch]{lineno}
\usepackage{cite}
\usepackage{float}

\def\BibTeX{{\rm B\kern-.05em{\sc i\kern-.025em b}\kern-.08em
    T\kern-.1667em\lower.7ex\hbox{E}\kern-.125emX}}
\usepackage{balance}
\usepackage{multirow}

\usepackage[colorlinks,
            linkcolor=blue,
            anchorcolor=blue,
            citecolor=blue]{hyperref}

\begin{document}

\title{SonarT165: A Large-scale Benchmark and STFTrack Framework for Acoustic Object Tracking}

\author{Yunfeng Li, Bo Wang*, Jiahao Wan, Xueyi Wu, Ye Li.
\thanks{This research is funded by the National Natural Science Foundation of China, grant number 52371350, by the National Key Research and Development Program of China, grant number 2023YFC2809104, and by the National Key Laboratory Foundation of Autonomous Marine Vehicle Technology, grant number 2024-HYHXQ-WDZC03. (Corresponding author: Bo Wang.)
} }

\maketitle

\input{sections/0_abstract}
\input{sections/0_keywords}
\input{sections/1_introduction}
\input{sections/2_related_work}
\input{sections/3_benchmark}
\input{sections/4_method}

\input{sections/5_experiments}
\input{sections/6_discussion}
\input{sections/7_conclusion}

\bibliography{references}
\bibliographystyle{IEEEtran}
\end{document}

%% file: sections/0_abstract.tex
\begin{abstract}
Underwater observation systems typically integrate optical cameras and imaging sonar systems. When underwater visibility is insufficient, only sonar systems can provide stable data, which necessitates exploration of the underwater acoustic object tracking (UAOT) task. Previous studies have explored traditional methods and Siamese networks for UAOT. However, the absence of a unified evaluation benchmark has significantly constrained the value of these methods. To alleviate this limitation, we propose the first large-scale UAOT benchmark, SonarT165, comprising 165 square sequences, 165 fan sequences, and 205K high-quality annotations. Experimental results demonstrate that SonarT165 reveals limitations in current state-of-the-art SOT trackers. To address these limitations, we propose STFTrack, an efficient framework for acoustic object tracking. It includes two novel modules, a multi-view template fusion module (MTFM) and an optimal trajectory correction module (OTCM). The MTFM module integrates multi-view feature of both the original image and the binary image of the dynamic template, and introduces a cross-attention-like layer to fuse the spatio-temporal target representations. The OTCM module introduces the acoustic-response-equivalent pixel property and proposes normalized pixel brightness response scores, thereby suppressing suboptimal matches caused by inaccurate Kalman filter prediction boxes. To further improve the model feature, STFTrack introduces a acoustic image enhancement method and a Frequency Enhancement Module (FEM) into its tracking pipeline. Comprehensive experiments show the proposed STFTrack achieves state-of-the-art performance on the proposed benchmark. The code is available at \url{https://github.com/LiYunfengLYF/SonarT165}.
\end{abstract}

%% file: sections/0_keywords.tex
\begin{IEEEkeywords}
Underwater Acoustic Object Tracking, Tracking Benchmark, Spatio-Temporal Fusion, Trajectory Prediction, Single Object Tracking.
\end{IEEEkeywords}

%% file: sections/1_introduction.tex
\section{Introduction}
Underwater optical cameras and imaging sonar systems serve as the primary sensing modalities for underwater observation \cite{rgbs50}. Due to severe light attenuation in underwater environments, the effective operational range and reliability of optical cameras degrade rapidly, whereas sonar systems leverage acoustic waves to achieve superior robustness and extended detection ranges. This performance gap implies that underwater vehicles equipped with both sensing modalities must prioritize acoustic data under low-visibility conditions (Figure \ref{fig: motivation}). Therefore, exploring underwater acoustic object tracking is critical to enhance the operational efficiency of underwater observation platforms.

\input{figs/motivation}

Underwater acoustic object tracking (UAOT) is a combination of single object tracking (SOT) and underwater acoustic vision task (sonar image processing), aiming to locate the position and scale of an acoustic target within sequential sonar frames. In contrast to optical imagery (e.g., RGB, thermal, or depth image), acoustic images are single-channel representations encoding acoustic back-scatter intensity (0-255 grayscale), with pixel values directly proportional to signal strength at corresponding spatial coordinates. Two inherent limitations distinguish the acoustic image: (1) low-texture regions resulting from sparse acoustic reflectors and (2) high background noise caused by multipath interference and turbulent flow. Furthermore, acoustic artifacts (morphologically similar to the true target) frequently arise from seabed reverberation and sidelobe effects. Overall, these issues pose challenges for the application of SOT trackers in acoustic object tracking.

Previous research on UAOT has explored various methods: Kalman filters \cite{sonartrackingkf1}\cite{sonartrackingkf2}, particle filters \cite{sonartrackingpf1}\cite{sonartrackingpf2}, machine learning techniques \cite{sonartrackingml1}\cite{sonartrackingml2}, Siamese-Network \cite{sonartrackingsiam1}\cite{sonartrackingsiam2}, and custom neural architectures \cite{sonartrackingdn1}\cite{sonartrackingdn2} to achieve the tracking task. Other studies \cite{sonartrackingdet1}\cite{sonartrackingdet2} have combined YOLO-style detectors and trajectory matching to track an acoustic target. However, the impact of these efforts has been limited by the absence of a standardized, large-scale benchmark dataset. Although the RGBS50 \cite{rgbs50} dataset offers some sonar test sequences, its limited size makes it difficult to promote the development of acoustic trackers. Overall, UAOT task is at a very early stage of research.

To alleviate these issues, we propose SonarT165, the first large-scale underwater acoustic object tracking benchmark, comprising 330 test sequences (165 square- and 165 fan-shaped) along with 205K high-quality annotations. All sequences are collected in pools and field environments to ensure their practicality. In addition, we evaluate state-of-the-art general trackers and lightweight trackers on the proposed benchmark. The experimental results show that the trackers achieve competitive results in precision rate (PR) scores, but their performance in success rate (SR) is insufficient. In general, SonarT165 presents a challenge to current SOT paradigm trackers.

Compared to objects in RGB images, acoustic objects have simpler contours, with high-intensity pixels in target regions and low-intensity backgrounds, resulting in well-defined edges. This contrast allows these trackers to achieve high precision rate (PR) scores. However, limitations in the principles of acoustic imaging \cite{rgbs50} cause the acoustic signature (target appearance) to vary drastically with the target position, resulting in an insufficient success rate (SR) performance. Furthermore, occlusion by other acoustic objects or their acoustic artifacts will merge the pixel of the target and interference into a large high-brightness region, making it difficult to distinguish targets based on appearance (acoustic) features.

Therefore, we propose a spatio-temporal trajectory fusion tracker family STFTrack for UAOT. STFTrack takes the LiteTrack \cite{litetrack} tracking pipeline as its baseline (LiteTrack-B8 \cite{litetrack} for STFTrack-B and LiteTrack-B6 \cite{litetrack} for STFTrack-S) and introduces an acoustic target enhancement method to enhance high frequency information of target appearance and a frequency enhancement module to improve target characteristic, respectively. STFTrack contains two novel modules: a multi-view template fusion module (MTFM) and an optimal trajectory correction module (OTCM). The MTFM module performs joint enhancement and cross-attention modeling on the original and binary images of dynamic targets, and then fuses multi-view dynamic templates with attention-based fixed templates. The OTCM module mitigates suboptimal matching caused by inaccurate Kalman filter predictions through pixel brightness response scores and intersection over box2 (IoB) scores derived from maximum response boxes. These metrics optimize the correct matching of target candidate boxes in the response map.

The main contributions are summarized as follows.
\begin{itemize}
\item[$\bullet$] We introduce the first large-scale UAOT benchmark. SonarT165, which contains 165 square sequences and 165 fan sequences, and 205K high-quality annotations. In addition, we evaluate popular general trackers and lightweight trackers on the benchmark to promote the development of acoustic object tracking.

\item[$\bullet$] We propose a novel Multi-view Template Fusion Module (MTFM), which generates multi-view dynamic templates using original and binary images, then fuses spatio-temporal target representations via fixed and dynamic templates.

\item[$\bullet$] We propose a novel optimal trajectory correction module (OTCM), which introduces a normalized brightness pixel response score of the target and an intersection over box2 (IOB) score of the maximum response box to mitigate the suboptimal matching of inaccurate Kalman boxes.

\item[$\bullet$] Comprehensive experiments demonstrate that the proposed STFTrack tracking pipeline achieves state-of-the-art performance among general trackers and lightweight trackers on the proposed SonarT165 benchmark.

\end{itemize}

%% file: figs/motivation.tex
\begin{figure}
	\centering
        \includegraphics[width=9cm]{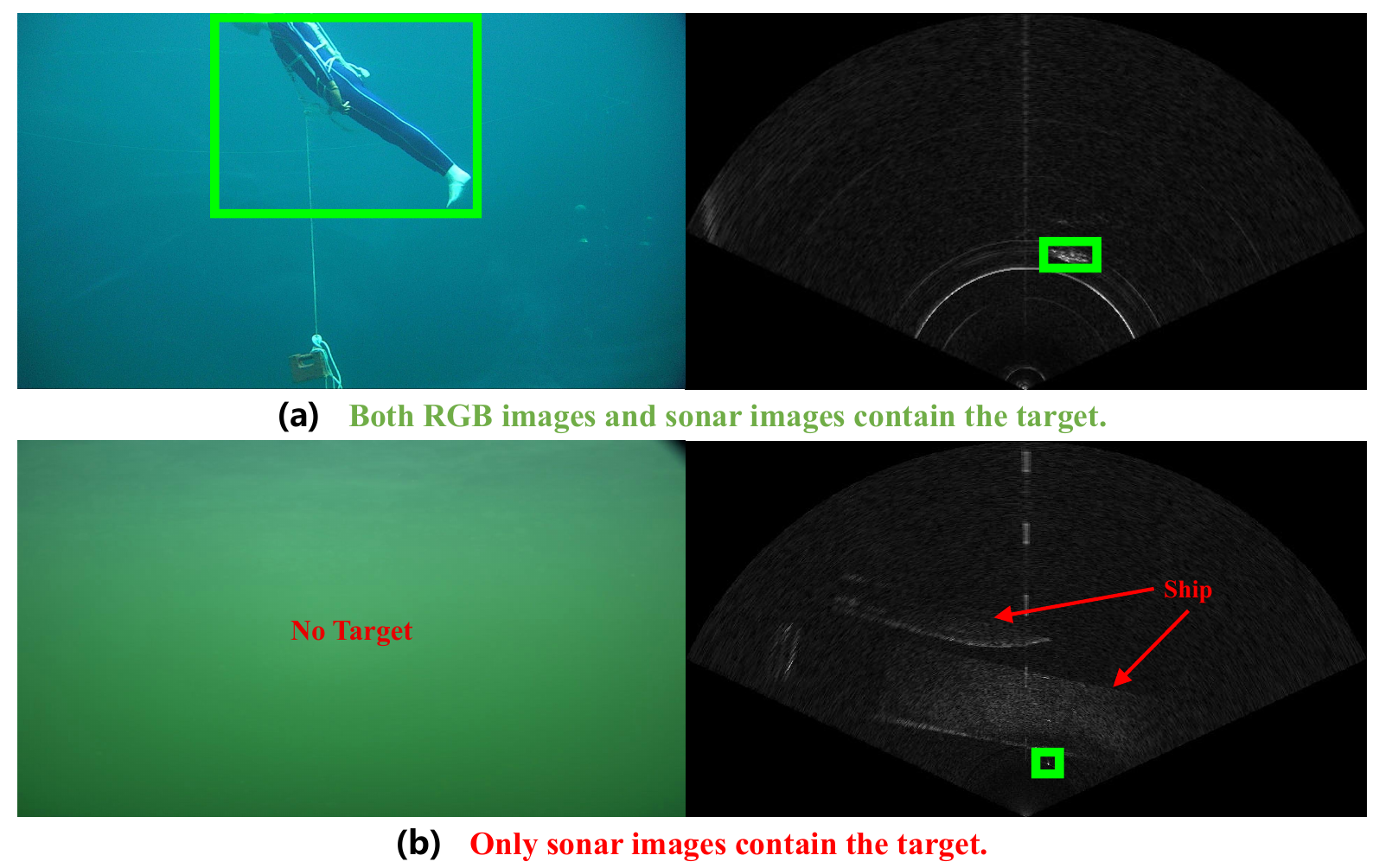}
	\caption{When underwater visibility is sufficient (in figure (a)), vehicle can use underwater camera and sonar system to jointly locate the tracked target, such as RGB-Sonar tracking \cite{rgbs50} task. When underwater visibility is insufficient (in figure (b)), vehicle needs to rely on sonar alone to locate the target, which is the underwater acoustic object tracking (UAOT) task. }
	\label{fig: motivation}
\end{figure}

%% file: sections/2_related_work.tex
\section{Related Work}
\subsection{Single Object Tracking}
Single object tracking (SOT) supports tracking of all types of target, making it directly applicable to UAOT task. Popular SOT trackers include Siamese trackers and Transformer trackers. The Siamese trackers \cite{siamrpn}\cite{siamrpn++}\cite{siamban}\cite{siamcar} model correlations by performing different correlation operations on template feature and search area feature. The Transformer trackers \cite{stark}\cite{ostrack}\cite{odtrack}\cite{lorat} achieve attention-based relationship modeling through attention-like networks. These methods show the main framework of SOT trackers. In addition, techniques such as spatio-temporal information utilization \cite{stark}\cite{tomp}\cite{hiptrack}\cite{tip_v1}\cite{tip_v2}, trajectory prediction fusion \cite{uostrack}\cite{neighbortrack}\cite{tip_t1}, sequence training methods \cite{transtslt}, feature enhancement \cite{aiatrack}\cite{lorat}, and better framework \cite{ostrack}\cite{seqtrack} are introduced into SOT trackers to improve model performance. Lightweight tracking is a lightweight implementation of SOT. Similarly, it can also be divided into the Siamese model \cite{lighttrack}\cite{fear}\cite{lightfc} and the Transformer model \cite{mixformerv2}\cite{hit}\cite{litetrack}, depending on the modeling approach.

Although these trackers can be directly applied to the UAOT task, experimental results indicate that our SonarT165 benchmark presents new challenges to these trackers.

\subsection{Underwater Acoustic Object Tracking}
The classic methods \cite{sonartrackingkf1}\cite{sonartrackingkf2}\cite{sonartrackingpf1}\cite{sonartrackingpf2} for UAOT are to use digital image processing methods to obtain the target position and to use a Kalman filter (or its combination with other filters) to track the target. Although they strongly contribute to the development of UAOT task, the lack of depth-feature-based discrimination makes it difficult for these methods to handle appearance variations. Some YOLO-based acoustic trackers \cite{sonartrackingdet1}\cite{sonartrackingdet2} also achieve the tracking task, but they also face the problems of high global computation consumption and identity switching. In addition, \cite{sonartrackingsiam1} employs a fully convolutional network, while \cite{sonartrackingsiam2} incorporates an attention mechanism to develop Siamese-based acoustic trackers, but their simple architectures are not sufficient to support complex acoustic object tracking scenarios. Overall, these works focus on using traditional and shallow features to achieve the tracking tasks. In comparison to these works, our work explores the combination of advanced trackers with sonar image features and acoustic tracking challenges.

Current UAOT task lacks large-scale tracking benchmarks, although in similar tasks such as RGB-Sonar tracking, the RGBS50 \cite{rgbs50} can provide a number of sonar test sequences to evaluate the tracker, but there are limitations to its size. Compared to RGBS50 \cite{rgbs50}, our SonarT165 benchmark has a larger scale (205k \textit{v.s.} 44K), more sequences (330 \textit{v.s.} 50), and richer scenarios (pool and field environments \textit{v.s.} pool environments). Overall, our benchmarks are more conducive to promoting the development of acoustic object tracking.

\subsection{Spatio-Temporal Template Fusion}

Spatio-temporal template fusion improves model discrimination of the target with appearance variations. Some template fusion methods \cite{stark}\cite{tatrack} integrate the fixed template, dynamic template, and search area through attention layers during the tracking process. These methods improve tracking performance at the cost of increased computational consumption. 
Some template fusion methods \cite{updatenet}\cite{fear}\cite{lightfcx} precompute the fused template, which interacts with the search area. UpdateNet \cite{updatenet} proposes a fully convolutional template update network. FEAR \cite{fear} explores a template fusion method based on cosine similarity. LightFC-X \cite{lightfcx} explores a dual-template joint modeling method through an attention layer.

Compared to them, our method combines traditional processing methods into an acoustic vision task, using both original images and binary images of the dynamic template to model multi-view feature, and then modeling the spatio-temporal representation of the target through two templates.

\subsection{Trajectory Prediction for Tracking}

Trajectory prediction method provides motion-based position priors and avoids tracking drift caused by incorrect appearance discrimination. Kalman filter \cite{tpm_kf1}\cite{tpm_kf2}, IMM \cite{tpm_imm1}, mean shift \cite{tpm_meanshift}, and other motion estimation methods \cite{tpm_other1}\cite{tpm_other2} are proposed to track satellite objects with relatively simple motion patterns. In addition, the response map encodes the target and other objects in the search area. NeighborTrack \cite{neighbortrack} models the trajectories of the target and other objects to deal with occlusion and similar appearance challenges for SOT. In the UOT task, UOSTrack \cite{uostrack} uses trajectory prediction boxes as priors and matches candidate boxes that satisfy motion priors within the response map. Similarly, ATCTrack \cite{atctrack} enhances UOSTrack \cite{uostrack} by replacing IoU with center-point distance metrics, better aligning with UOT motion patterns.

Compared with them, our method utilizes the characteristic of acoustic image sound reflection intensity equal to pixel value to mitigate the suboptimal bounding box matching caused by inaccurate prediction of Kalman filter.

%% file: sections/3_benchmark.tex
\section{Sonar Tracking Benchmark}
\input{figs/benchmark_overall}
\input{figs/benchmark_char}
\input{figs/benchmark_distribution}

\subsection{SonarT165 Benchmark}
We collect 165 underwater acoustic video sequences and processed them in both raw and fan image formats to obtain a total of 330 test sequences for evaluation. Among 165 videos, 117 are collected in a pool, while the remaining 48 are collected in a wild environment. All annotations are manually annotated and each annotation is proofread by a full-time annotator to ensure consistency in the description of the target appearance by the bounding box. We provide more SonarT165 benchmark details as follows:

\subsubsection{Hardware Setup} 
We use Oculus MD750 sonar to collect data. It is installed on a sensor platform in the pool and on an AUV in the field environment. The sonar used operates in high-frequency mode and samples at a speed of 10 \textit{fps}. The depth of the pool is 10 meters and the target is suspended and dragged at a distance of 3-7 meters from the water surface. The wild environment is located in the Danjiangkou Reservoir, Danjiangkou City, Henan Province, China. The target category and motion settings are the same as for the pool.

\subsubsection{Annotation}

We manually annotate the target bounding box in the format of $[X, Y, W, H] $, where $X$ and $Y$ represent the coordinates of the upper left corner point, and $W$ and $H$ represent the width and height, respectively. The box is annotated as $[0,0,0,0]$ when the target is out-of-view. Due to the principle of acoustic imaging, the sound reflection intensity of the target decreases at a long distance, resulting in low pixel values and making the target partially invisible in the acoustic image. Therefore, all targets are annotated only in the visible part.

\subsubsection{Statistics} 

We analyze the statistics of our SonarT165 benchmark as follows:

\begin{itemize}
\item[$\bullet$]\textit{Benchmark Scale:}
Our SonarT165 benchmark includes 165 square image test sequences and 165 fan image test sequences, totaling 330 sequences and 205,288 frames. The minimum frame number, average frame number, and maximum frame number of the test sequences are 62, 622, and 3,356, respectively.

\item[$\bullet$]\textit{Attributes:} Our SonarT165 benchmark contains 10 different attributes: Acoustic Object Crossover (AOC), Similar Object (SO), Out-of-View (OV), Small Target (ST), Scale Variant (SV), Appearance Change (AC),  Low Acoustic Reflection (LAR), Target Brightness Change (TBC), Background Interference (BI), Field Environment (FE). We provide detailed definitions of these attributes in Table \ref{table benchmark attribute}. The frame and sequence level distribution of each attribute is shown in Figure \ref{fig: benckmark} (c) and (f). The visualization of the attributes is shown in Figure \ref{fig: attributes}.

\item[$\bullet$]\textit{Object Categories:}
Our category settings follow the RGBS50 \cite{rgbs50} benchmark and include a total of 7 categories, which are ball and polyhedron, connected polyhedron, fake person, frustum, iron ball, octahedron, and UUV (includes 2 different sizes).

\item[$\bullet$]\textit{Box Distribution:}
We present the box distribution of SonarT165 benchmark in Figure \ref{fig: box distribution}.
The initial frame box and all box images have distribution in all regions. In addition, the average square root of the width times height of the boxes in most of our sequences is around 20, which means that our target size is relatively small.

\item[$\bullet$]\textit{Two Types of Sequences:} SonarT165 includes two typical acoustic sequences: square sequence and fan sequence, as shown in Figure \ref{fig: box distribution}. Two types of sequences will help trackers adapt to different acoustic image formats.

\end{itemize}

To our knowledge, our proposed SonarT165 is the first large-scale UAOT benchmark dataset.

\subsection{Comparison with Other Tracking Benchmark Datasets}
We compare the proposed SonarT165 dataset with other tracking benchmark datasets, as shown in Table \ref{table benchmark compare}.

\subsubsection{Benchmark Scale}

As a combination of SOT and acoustic vision task, UAOT task currently lacks large-scale tracking benchmarks. Our SonarT165 aims to alleviate this issue. Compared to the SOT benchmarks, the scale of SonarT165 is 3.5x larger than OTB100 \cite{otb15} and 1.8x larger than UAV123 \cite{uav123}. Compared to the UOT benchmarks, it is 2.8x larger than UOT00 \cite{uot100} and \cite{vmat}, 3.5x larger than UTB180 \cite{utb180}. Compared to RGBS50 \cite{rgbs50}, which is the most similar dataset to ours, the number of acoustic sequences and frames is 6.6x and 4.7x higher, respectively. 

\input{tables/benchmark_compare}
\input{tables/benchmark_attribute}

\subsubsection{Image Differences}
The images that the tracker needs to process have significant differences in different tracking tasks. In the SOT task, images are usually typical open-air images with rich backgrounds and targets. In the UOT task, underwater images typically exhibit image degradation and color distortion. Compared to them, acoustic images consist of intensity maps of sound reflections within a region, with a single (black) background and significant background noise (salt-and-pepper noise in the image). The target is a grayscale object composed of reflections from different positions of itself. When the target position changes, the reflection intensity of each part will change, resulting in the rotation and deformation, etc. of the object in the image. In addition, changes in the distance of the target also cause changes in the intensity of sound reflection, resulting in changes in the brightness (pixel value) of the object in the image. Therefore, UAOT task naturally need to deal with a series of problems such as strong background noise, weak target texture, appearance changes, and brightness changes.

\subsection{Baseline Methods}
In order to comprehensively evaluate the performance of current popular SOT trackers in the UAOT task, we select Siamese trackers, online-discriminator trackers and Transformer trackers as baselines to evaluate their performance on the proposed SonarT165 benchmark. In addition, considering that acoustic target trackers may be deployed on water downloading gear, we similarly select popular lightweight trackers and evaluate their performance. For ease of differentiation, we refer to non-lightweight trackers as general trackers.

The general baseline trackers innclude SiamRPN \cite{siamban}, SiamRPN++ \cite{siamrpn++}, DiMP18 \cite{dimp}, DiMP50 \cite{dimp}, PrDiMP18 \cite{prdimp}, PrDiMP50 \cite{prdimp}, SiamCAR \cite{siamcar}, SiamBAN \cite{siamban}, SiamBAN-ACM \cite{siambanacm}, KeepTrack \cite{keeptrack}, TrDiMP50 \cite{trdimp}, StarkS50 \cite{stark}, StarkST50 \cite{stark}, StarkST101 \cite{stark}, ToMP50 \cite{tomp}, ToMP101 \cite{tomp}, OSTrack256 \cite{ostrack}, OSTrack384 \cite{ostrack}, AiATrack \cite{aiatrack}, UOSTrack \cite{uostrack}, ARTrackSeq-B256 \cite{artrack}, SeqTrack-B256 \cite{seqtrack}, SeqTrack-B384 \cite{seqtrack}, SeqTrack-L256 \cite{seqtrack}, SeqTrack-L384 \cite{seqtrack}, HiPTrack \cite{hiptrack}, ODTrack-B256 \cite{odtrack}, ODTrack-L256 \cite{odtrack}, ARTrackV2Seq-B256 \cite{artrackv2}, LoRAT-B224 \cite{lorat}, LoRAT-B378 \cite{lorat}, LoRAT-L224 \cite{lorat}, LoRAT-L378 \cite{lorat}, LoRAT-G224 \cite{lorat}, LoRAT-G378 \cite{lorat}, MCITrack-B224 \cite{mcitrack}, MCITrack-L224 \cite{mcitrack}, MCITrack-L384 \cite{mcitrack}. 

The lightweight baseline trackers include MobileSiamRPN++ \cite{siamrpn++}, HiT \cite{hit}, LightFC \cite{lightfc}, LightFC-vit \cite{lightfc}, SMAT \cite{smat}, LiteTrack-B4 \cite{litetrack}, LiteTrack-B6 \cite{litetrack}, LiteTrack-B8\cite{litetrack}, LiteTrack-B9 \cite{litetrack}, MCITrack-T224 \cite{mcitrack}, MCITrack-S224 \cite{mcitrack},

Overall, the above trackers reflect the advanced technology and latest progress of SOT task, and introducing them into the SonarT165 benchmark can promote the development of UAOT task.

\subsection{Evaluation Metrics}

We follow the One-Pass Evaluation Protocol (OPE) to evaluate the baseline trackers. We follow the widely used metrics PR, NPR, and SR in the tracking community to evaluate the tracker. In addition, we introduce OP50, OP75, and F1 scores to describe the tracking ability at medium-precision, high-precision tracking ability, and recognition ability for target positive samples, respectively.

\begin{itemize}
\item[$\bullet$]\textbf{Precision Rate (PR)}. We calculate the PR score through the percentage of frames where the distance between the predicted position and the ground truth is within a threshold of 20.

\item[$\bullet$]\textbf{Normalized Precision Rate (NPR)}. Following the setting of \cite{rgbs50}, the NPR score is introduced to eliminate the impact of image size and box size on accuracy.

\item[$\bullet$]\textbf{Success Rate (SR)}. We first obtain the success rate curve by calculating the overlap rate between the ground true and predicted boxes that is greater than different thresholds. Then we obtain the SR score through the area under the curve.

\item[$\bullet$]\textbf{Overlap Precision at 50\% (OP50)}. We calculate the proportion of frames with an Intersection over Union (IoU) of more than 50\% between predicted boxes and ground truth.

\item[$\bullet$]\textbf{Overlap Precision at 75\% (OP75)}. We calculate the proportion of frames with an IoU of more than 75\% between predicted boxes and ground truth.

\item[$\bullet$]\textbf{F1 Score (F1)}. We first count the True Positives (TP), False Positives (FP), and False Negatives (FN). We then calculate the Precision (P) and Recall (R) by $P=\frac{TP}{TP+FP}$ and $R=\frac{TP}{TP+FN}$. Finally, we calculate the F1 score by $F1=\frac{2\times P\times R}{P+R}$.

\end{itemize}

Because of the differences in target appearance between square sonar images and fan images, we propose to evaluate the two types of image sequences separately.

%% file: figs/benchmark_overall.tex
\begin{figure*}
	\centering
        \includegraphics[width=18cm]{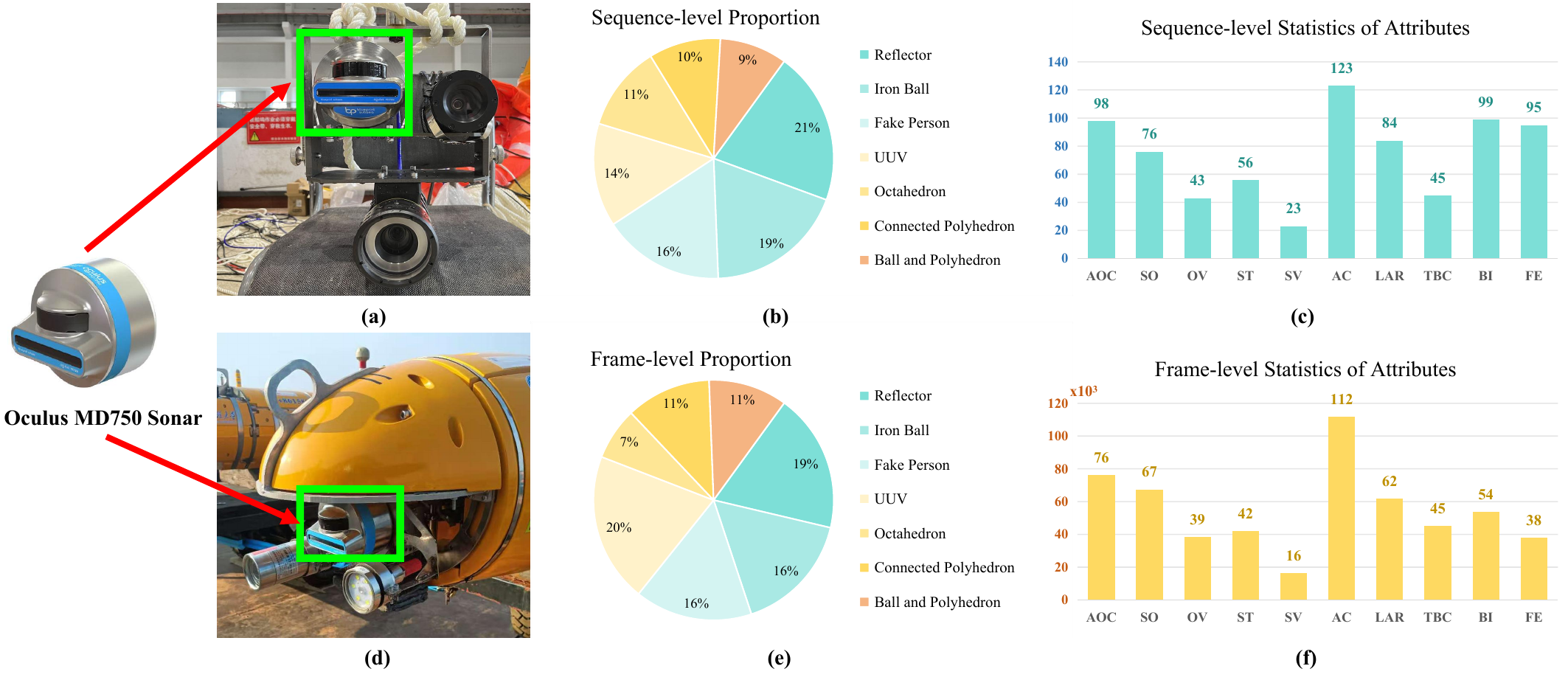}
	\caption{Main introduction of the proposed SonarT165 benchmark. (a) Data collection platform in the pool. (b) Sequence level proportion of different objects. (c) Sequence level statistics of different attributes. (d) Data collection platform in the field environment. (e) Frame level proportion of different objects. (f) Frame level statistics of different attributes.}
	\label{fig: benckmark}
\end{figure*}

%% file: figs/benchmark_char.tex
\begin{figure*}
	\centering
        \includegraphics[width=18cm]{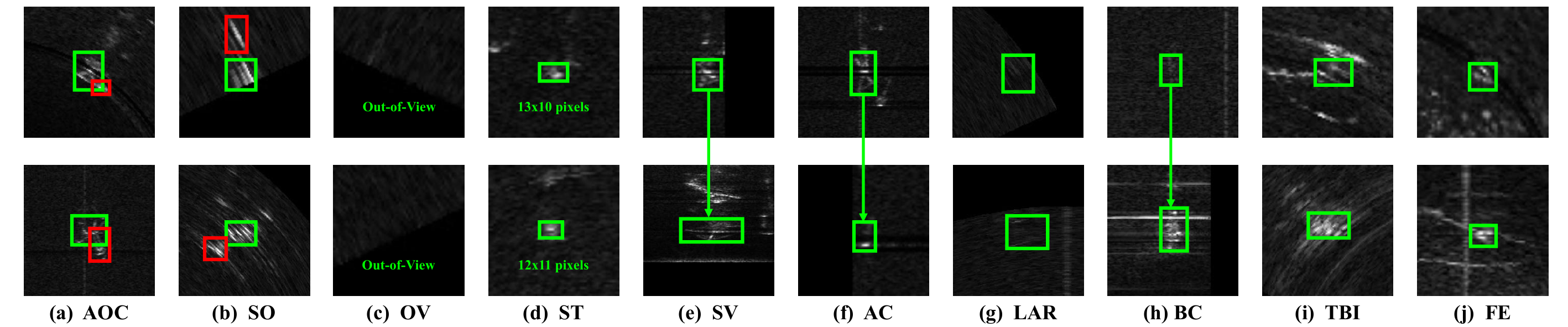}
	\caption{Visualization of different attributes of the proposed SonarT165 benchmark. To show more intuitively the challenges they pose to the tracker, we show them in the search area. (a) Acoustic object crossover . (b) Similar object. (c) out-of-view. (d) Small target. (e) Scale variant. (f) Appearance change. (g) Low acoustic reflection. (h) Target brightness change. (i) Background interference. (j) Field environment.}
	\label{fig: attributes}
\end{figure*}


%% file: figs/benchmark_distribution.tex
\begin{figure*}
	\centering
        \includegraphics[width=18cm]{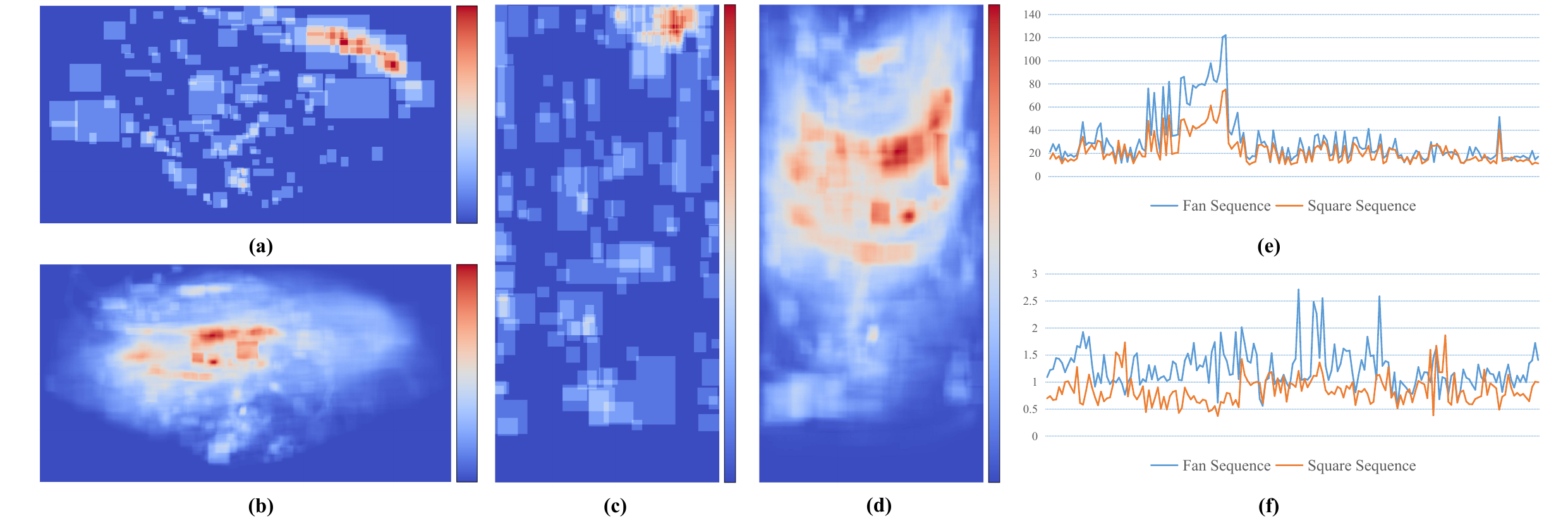}
	\caption{Visualization of bounding box distribution. (a) represents the distribution of the first frame bounding box in the fan sequences. (b) represents the distribution of all bounding boxes in the fan sequences. (c) represents the distribution of the first frame bounding box in the square sequences. (b) represents the distribution of all bounding boxes in the square sequences. (e) represents the square root curve of the width and height of bounding boxes in two types of sequences. (f) represents the width-height ratio curve of bounding boxes in two types of sequences.}
	\label{fig: box distribution}
\end{figure*}

%% file: tables/benchmark_compare.tex
\begin{table}[]
\caption{Compare with UAOT benchmark and other task benchmarks.}
\label{table benchmark compare}
\centering
\renewcommand{\arraystretch}{1.1}
\setlength{\tabcolsep}{1.2pt}

\begin{tabular}{c|c|cccccc}
\hline
Task                 & Benchmark & \thead{Num. \\Classes} & \thead{Num. \\ Seq} & \thead{Min. \\ Frames} & \thead{Avg. \\ Frames} & \thead{Max. \\ Frames} & \thead{Total. \\ Frames} \\ \hline
\multirow{3}{*}{SOT} & OTB15 \cite{otb15}    & 16 & 100  & 71         & 590        & 3,872      & 59K          \\
                     & TC128 \cite{tc128}    & 27 & 128 & 71         & 429        & 3,872      & 55K          \\
                     & UAV123 \cite{uav123}  & 9  & 123 & 109        & 915        & 3,085      & 113K         \\ \hline
\multirow{4}{*}{UOT} & UOT100 \cite{uot100}  & -  & 106     & 264        & 702        & 1,764      & 74K          \\
                     & UTB180 \cite{utb180}  & -  & 180     & 40         & 338        & 1,226      & 58K          \\
                     & VMAT \cite{vmat}      & 17 & 33      & 438        & 2,242      & 5,550      & 74K          \\
                     & UVOT400 \cite{uvot400}& 50 & 400     & 40         & 688        & 3,273      & 275K         \\ \hline
RGB-S                & RGBS50 \cite{rgbs50}  & 7  & 50      & 251        & 874        & 2,740      & 44K          \\ \hline
UAOT      & SonarT165             & 7  & 300     & 62         & 622        & 3,356      & 205K         \\ \hline
\end{tabular}
\end{table}

%% file: tables/benchmark_attribute.tex
\begin{table}[]
\caption{List and description of 10 attributes.}
\label{table benchmark attribute}
\renewcommand{\arraystretch}{1.1}
\centering
\setlength{\tabcolsep}{2.5pt}
\begin{tabular}{l|l}
\hline
\textbf{Attri.} & \textbf{Definition} \\ \hline
\textbf{AOC}        & Acoustic Object Crossover -The target overlaps with the position  \\
                    & or acoustic ghosting of another object.                           \\
\textbf{SO}         & Similar Object - The target is surrounded by similar appearance   \\
                    & objects.                                                          \\ 
\textbf{OV}         & Out-of-View - The target moves out of view and returns.           \\
\textbf{ST}         & Small Target - The target width and height are both less than 15. \\
\textbf{SV}         & Scale Variant - The scale change rate of the bounding box exceeds \\
                    & the range of [0.5, 2].                                            \\
\textbf{AC}         & Appearance Change - The appearance of the target has significant  \\ 
                    & changes. It is regarded as a collection of deformation, rotation, \\
                    & and partial out-of-view                                           \\
\textbf{LAR}        & Low Acoustic Reflection - the target has a low brightness value   \\
                    & in acoustic images.                                               \\
\textbf{TBC}         & Target Brightness Change - The pixel brightness value of the      \\
                    & target shows shows significant changes.                           \\
\textbf{BI}         & Background Interference - Background noise or acoustic black      \\ 
                    & lines interfere with the prediction of the target.                \\
\textbf{FE}         & Field Environment - Reflect the acoustic characteristics of the   \\
                    & target in a lake environment.                                     \\ \hline
\end{tabular}
\end{table}

%% file: sections/4_method.tex
\section{Method}
\input{figs/framework}
\input{figs/method_sonar_enhancement}
\input{figs/method_freq_enhance_module}

The overall framework of STFTrack is illustrated in Figure \ref{fig: framework}. It contains a acoustic image enhancement method for improving image quality, a backbone for asynchronous feature extraction and modeling, and a frequency enhancement module for decoupled high- and low- frequency feature learning, which form the basic tracking pipeline. The proposed template fusion module and the trajectory correction module are inserted into the tracking pipeline. In addition, we train the baseline model and the template fusion module using grayscale images and RGBT images, respectively.

\subsection{Tracking Pipeline}
Firstly, we show the STFTrack tracking pipeline. Due to the characteristics of acoustic images, we combine acoustic image enhancement and feature frequency enhancement to improve the baseline pipeline \cite{litetrack}.

\textbf{Acoustic Image Enhancement.}
Acoustic (sonar) images typically contain background noise, and when the acoustic reflection intensity of the target is low, it is difficult to distinguish from the background due to the decrease in pixel values. This issue was overlooked in previous research. In addition, the brightness of sonar images reflects the acoustic reflection value of the area, which means that high-frequency information enhancement can better represent the reflection area of the target in the image. Therefore, we propose a high-frequency enhancement method for sonar images.

As shown in Figure \ref{fig: sonar image enhancement}, we first use Gaussian blur to extract the low-frequency image, then subtract the original image from the low-frequency image to obtain the high-frequency image, and finally add the high-frequency image twice to the original image to obtain the enhanced high-frequency image.

\textbf{Backbone.}
We take LiteTrack \cite{litetrack} as our baseline for asynchronous feature extraction and modeling. Firstly, the template and search area are represented as $z\in R^{3\times h_{z}\times w_{z}}$ and $x \in R^{3\times h_{x}\times w_{x}}$, respectively. They are embedded into $Z\in R^{C\times H_{z}\times W_{z}}$ and $X\in R^{C\times H_{x}\times W_{x}}$, where $H_{i},W_{i} =h_{i}/16,w_{i}/16,i\in \{z,x\}$. The asynchronous feature extraction process of templates and search areas is represented as:
\begin{equation}
\begin{split}
Attn_{z}^{n} & = \text{softmax}(Q_{z}K_{z}^T)V_{z} \\
Attn_{x}^{m} & = \text{softmax}(Q_{x}K_{x}^T)V_{x}
\end{split}
\end{equation}
where $Q$, $K$, and $V$ is the Query, Key, and Value matrices. $Attn$ represents the attention layer. $m$ and $n$ represent the number of layers, and $n>m$.

The modeling process of the relationship between the template and search area is represented as:
\begin{equation}
\begin{split}
Attn_{xz}^{n-m} & = \text{softmax}(Q_{x}[K_{x};K_{z}]^T)[V_{x}; V_{z}] \\
                & \triangleq [Z_{template};X_{search}]
\end{split}
\end{equation}
where $X_{search}$ is the output feature of the backbone.

\input{figs/method_template_fusion}
\input{figs/method_traj_model}

\textbf{Frequency Enhancement.}
We propose a frequency enhancement module (FEM). The FEM module improves the representation of search area feature by decoupling the learning of high-frequency and low-frequency features, as shown in Figure \ref{fig: freq enhance module}.

High-frequency feature enhancement aims to improve the texture and contour features of the target. It is implemented by an unbiased and learnable Laplacian convolution kernel, represented as:
\begin{equation}
X_{high} = \alpha\times\text{Conv}_{h}(X_{search})
\end{equation}
where $\alpha$ is a learnable parameter and is initialized to $1$. $\text{Conv}_{3\times 3}^{'}$ updates weights during training and is initialized to:
\begin{equation}
\text{Conv}_{h}^{init} =\begin{bmatrix}
-1 & -1 & -1 \\
-1 & \ 8 & -1 \\
-1 & -1 & -1
\end{bmatrix}
\end{equation}

Dynamic low-frequency feature enhancement aims to improve the smooth region features of the target. It is implemented by a dynamic Gaussian convolution kernel generated by a learnable parameter $\sigma$, represented as:
\begin{equation}
\begin{split}
\text{Conv}_{l}^{init} & = \text{GaussianKernel}(\sigma,\ ksize=5) \\
    X_{low} &=\text{Conv}_{l}(X_{search})
\end{split}
\end{equation}
where $\sigma$ is a learnable parameter and is initialized to $1$.

The output feature of the FEM module is represented as:
\begin{equation}
X_{fem} =X_{search}+X_{high}+X_{low}
\end{equation}
where $X_{fem}$ is fed into the prediction head.

\textbf{Head and Loss.} Following the design of LiteTrack \cite{litetrack}, we use a fully convolutional prediction head \cite{centerhead} to predict the target state and introduce weight focal loss \cite{weightfocalloss}, l1 loss, and GIoU \cite{giou} loss to train the model. The total loss is represented as:

\begin{equation}
L_{total}= L_{cls} + \lambda_{iou}L_{iou} + \lambda_{l_{1}}L_{1}
\end{equation}
where $\lambda_{iou}=2$ and $\lambda_{l_{1}}=5$ as same in \cite{litetrack}.

\subsection{Spatio-Temporal Template Fusion}
We propose a multi-view template fusion module (MTFM), which models the appearance representation of the target at different temporal states using multi-view images of acoustic images, as shown in Figure \ref{fig: template fusion}.

We first model the multi-view appearance of the dynamic template. The dynamic template is represented as $z_{d}\in R^{3\times h_{z}\times w_{z}}$. We combine the characteristics of pixel values reflecting acoustic reflection intensity to obtain acoustic multi-view images of the target.
\begin{equation}
z_{db}=\text{binary}(z_{d},\ thres=30)
\end{equation}
where $\text{binary}$ is the binarization operation.

We extract features of $z_{d}$ and $z_{db}$ and obtain $Z_{d}$ and $Z_{db}$. Then we perform multi-view spatial and channel enhancement on them separately, represented as
\begin{equation}
\begin{split}
Z_{d}^{f}  & = \text{Conv}_{1\times 1}(\text{concat}(Z_{d},\ Z_{db}) \\
Z_{d}^{ce} & = \text{Conv}_{1\times 1}(\text{Pooling}_{channel}(Z_{d}^{f})\times Z_{d}^{f}  \\
Z_{d}^{se} & = \text{MLP}(\text{Pooling}_{spatial}(Z_{d}^{f})\times Z_{d}^{f} \\
Z_{d}^{cs} & = Z_{d}^{ce}+Z_{d}^{se}
\end{split}
\end{equation}
where $\text{MLP}$ is a multi-layer perceptrons (MLP).

We then introduce cross-attention to model the multi-view appearance representation of the dynamic template.
\begin{equation}
Z_{mv}=\text{CrossAttn}(Z_{d},\ Z_{d}^{cs})+Z_{d}
\end{equation}
where $\text{CrossAttn}$ is a cross-attention layer.

Finally, we model the temporal representation of the target using the fixed template and the multi-view dynamic template, represented as:
\begin{equation}
\begin{split}
Z_{cross}  & = \text{CrossAttn}(Z,\ Z_{mv})+Z_{d}+Z_{mv} \\
Z_{fused}  & = \text{Linear}(Z_{cross})+Z_{d}+Z_{mv}
\end{split}
\end{equation}
where $Z_{fused}$ is the fused template.

The MTFM module integrates fixed templates, dynamic templates, and binary dynamic templates into a fusion template. It can avoid the impact of tracking inference on efficiency through pre-calculation during template updates.

\subsection{Trajectory Fusion}

UOSTrack \cite{uostrack} combines the Kalman filter and the reuse of the candidate box from the response map to improve the drift of the target tracking. However, the Kalman filter itself cannot provide an accurate box, as shown in Figure \ref{fig: tarj module} (a). Even if the gt boxes are used to update the filter, the average Iou of the predicted box is only about 0.8. Inaccurate Kalman prediction boxes result in suboptimal matching, as shown in Figure \ref{fig: tarj module} (b), the lagged Kalman prediction box may have better matching scores with the suboptimal boxes around the optimal box, but these suboptimal boxes are not as accurate as the optimal box, resulting in reduced accuracy, and error accumulation also leads to tracking drift of the target.

To alleviate this limitation, we propose an optimal trajectory correction module. It takes the UOSTrack \cite{uostrack} as a baseline and achieves the elimination of suboptimal matching based on the characteristics of the acoustic images. First, the response map predicted by the head is represented as $M\in R^{H_{x}W_{x}}$. We select the top-$k$ scores $S\in R^{k}$ and their candidate boxes $B_{c}\in R^{k}$. The predicted Kalman box is represented as $B_{kf}\in R^{1}$.

The IoU score $I_{box}$ that reflect the previous trajectory prior is represented as
\begin{equation}
I_{box} = \text{IoU}(B_{kf}, B_{c})\times S
\end{equation}

The acoustic target area and the background area are distinguished by the acoustic reflection values. More accurate bounding boxes typically cover high reflection areas of the target, thus containing higher pixel values. Therefore, we calculate the mean pixel response $R_{np}$ as
\begin{equation}
\begin{split}
x_{m} & = \text{binary}(x,\ thres)/255. \in R^{h_{x}w_{x}} \\
R_{np} & = \text{mean}(\text{extract\_patch}(x_{m},\ B_{c})) \in R^{k}
\end{split}
\end{equation}
where $\text{binary}$ represents binary segmentation of an image, $thres$ represents the segmentation threshold, which is obtained by calculating the average pixel value of the target in the previous frame.

Then we calculate the maximum score of $I_{box}\times R_{np}$ and select the matched bounding box $B_{m}$.

In addition, we introduce the intersection over box2 (IoB) score $I_{m}$ of $B_{m}$ and $B_{mr}$ to suppress the suboptimal bounding box around the maximum response value.
\begin{equation}
I_{m} = \text{IoB}(B_{m}, B_{mr})
\end{equation}
where $B_{mr}$ represents the box of the max response value of $M\in R^{H_{x}W_{x}}$. If $I_{m}$ is larger than 0.6, we consider $B_{m}$ as a suboptimal box and output $B_{mr}$; otherwise we output $B_{m}$.

%% file: figs/framework.tex
\begin{figure*}[t]
\centering
\includegraphics[width=18cm]{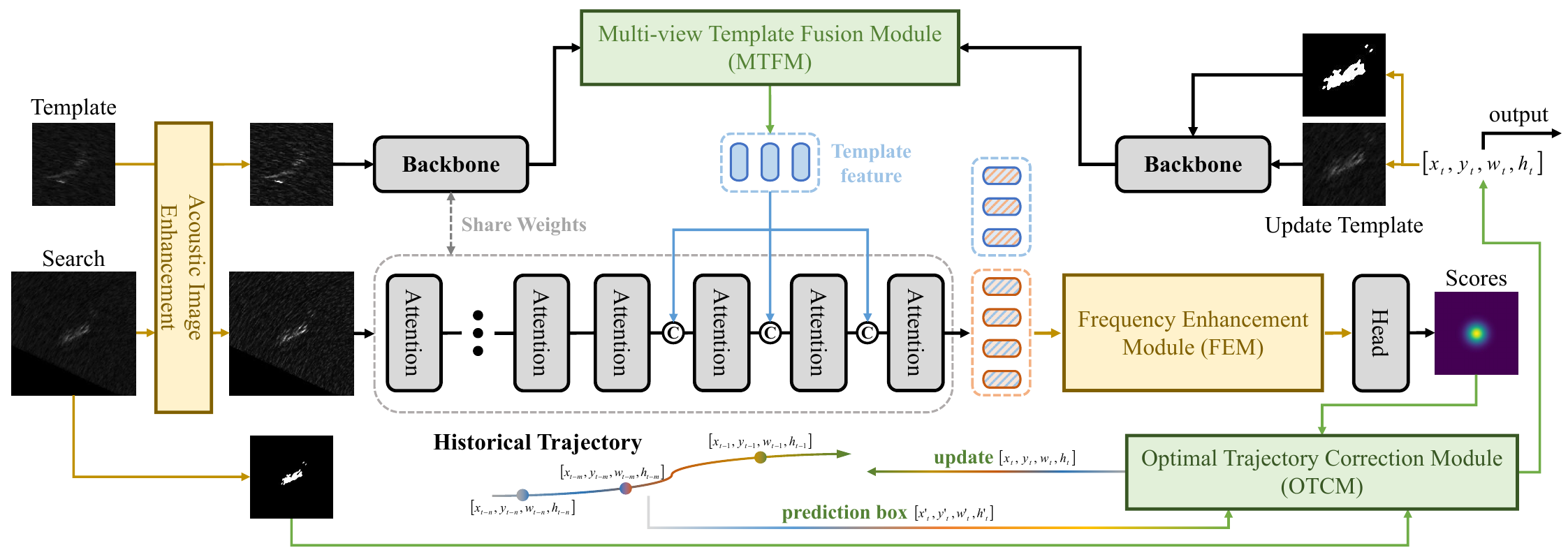}
\caption{The overall framework of STFTrack. We take SOT pre-trained Litetrack\cite{litetrack} as baseline. During the tracking phase, we first enhance the high frequency information of the sonar image and then encode the image and input it into the backbone. Then the search area features are input into the frequency enhancement module and then into the prediction head to obtain the target state. Then the predicted target state, target history state and acoustic response map are input into the trajectory correction module and output the bracketing frame. Then, we use the current frame bracket to obtain the dynamic template, and input the fixed template and dynamic template into the template fusion module to fuse the template features.}
\label{fig: framework}
\end{figure*}

%% file: figs/method_sonar_enhancement.tex
\begin{figure}[]
	\centering
    \includegraphics[width=7.5cm]{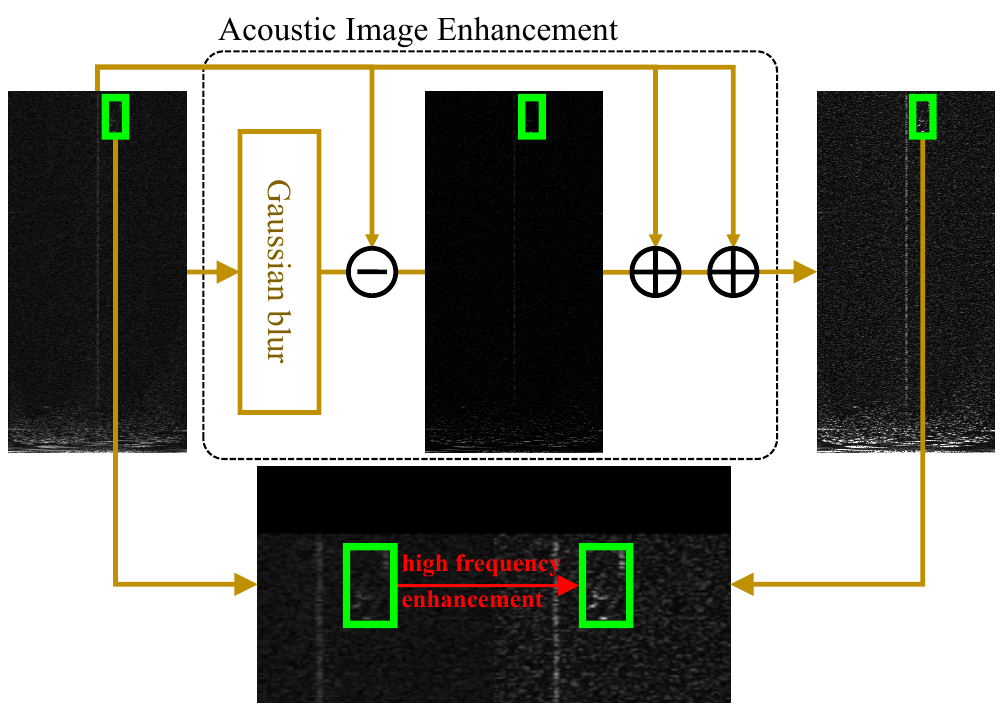}
	\caption{Presentation of acoustic image high-frequency enhancement.}
	\label{fig: sonar image enhancement}
\end{figure}

%% file: figs/method_freq_enhance_module.tex
\begin{figure}[]
	\centering
    \includegraphics[width=7cm]{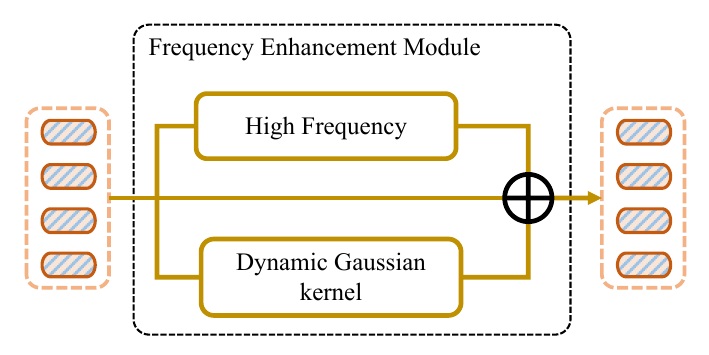}
	\caption{Presentation of frequency enhancement module.}
	\label{fig: freq enhance module}
\end{figure}

%% file: figs/method_template_fusion.tex
\begin{figure}[]
	\centering
    \includegraphics[width=8.6cm]{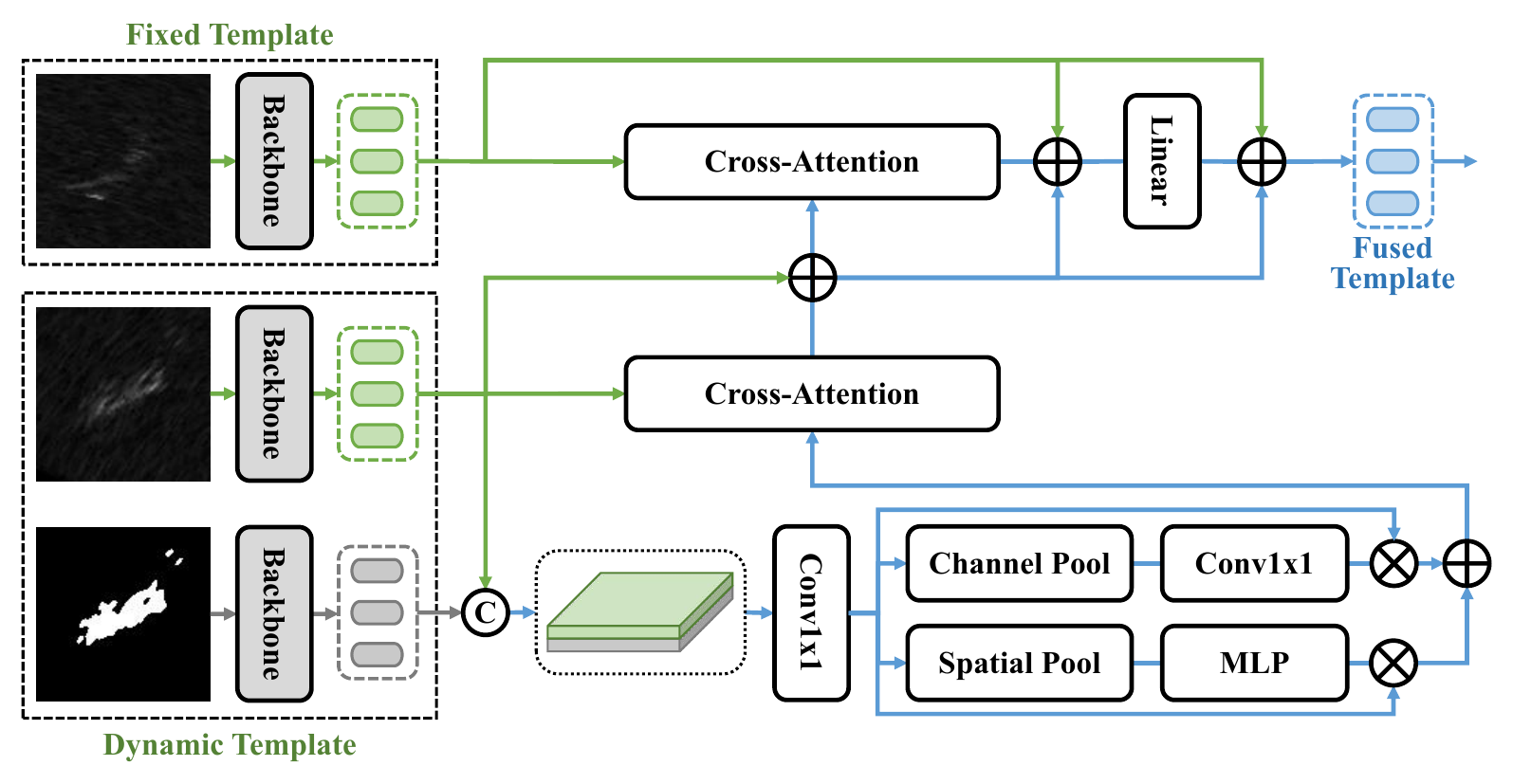}
	\caption{Presentation of the proposed multi-view template fusion module (MTFM). The dynamic template includes both the original image and the binary image.}
	\label{fig: template fusion}
\end{figure}

%% file: figs/method_traj_model.tex
\begin{figure}[]
	\centering
    \includegraphics[width=8.6cm]{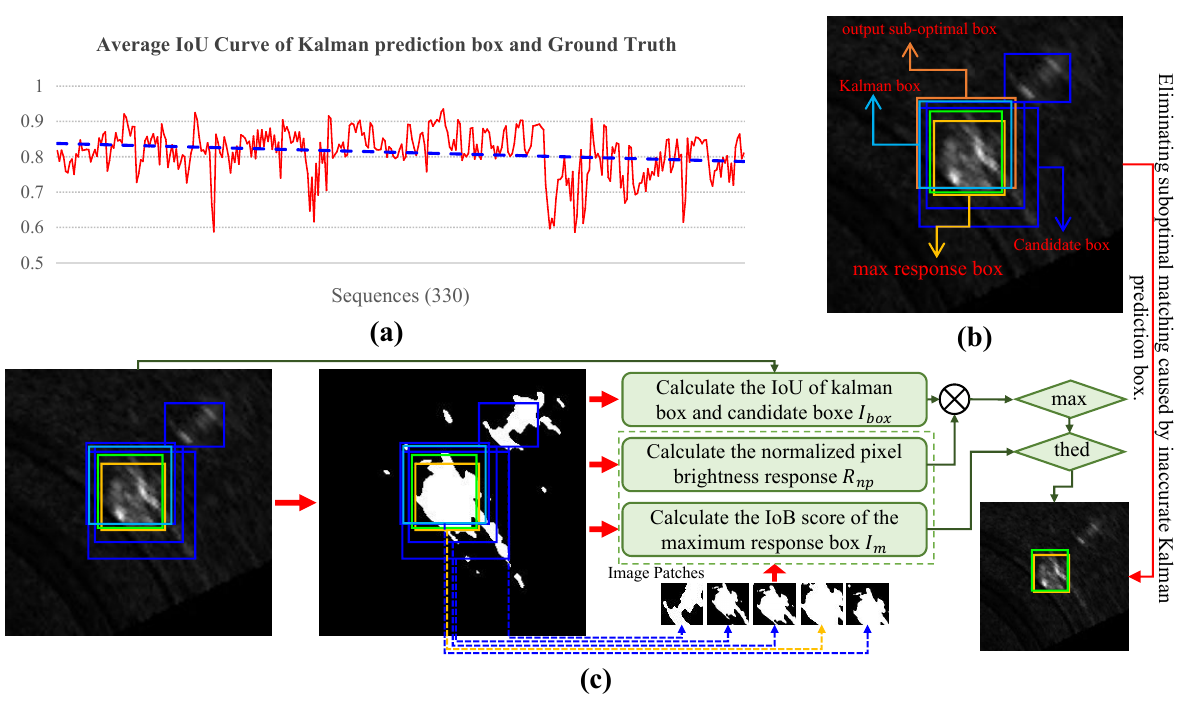}
	\caption{Presentation of the proposed optimal trajectory correction module (OTCM). (a) represents the Iou curve between the Kalman prediction box and the Ground truth, where the Kalman filter is updated by the Ground truth. (b) represents how the Kalman prediction box leads to suboptimal bounding box matching. (c) represents TCM module uses Kalman filter to eliminate candidate object interference while maintaining the optimal matching of the box.}
	\label{fig: tarj module}
\end{figure}

%% file: sections/5_experiments.tex
\section{Experiments}
\input{tables/baselines_results}
\input{figs/exp_plots}
\input{figs/exp_attr_radamaps}

\subsection{Implementation Details}
Our methods are implemented using Python 2.4.0 and Python 3.10. The training platform includes 2 Nvidia RTX A6000 GPUs. The training consists of two stages. The shared settings between the two stages are reported as follows. In each epoch, our sample number is 60000, and the total batch size is 64. The optimizer used is AdamW \cite{adamw} with a weight decay of $1 \times 10^{-4}$. The size of the template and the search area are $128\times 128$ and $256\times 256$, respectively.

\textbf{First Stage Training.} We train the Backbone, FEM module, and prediction head. The training set contains LaSOT \cite{lasot}, GOT10k \cite{got10k}, and UATD \cite{uatd}. During training, all RGB images are converted to grayscale images. The training epoch number is 10, which takes about 3 hours. The total learning rate is $1 \times 10^{-4}$. We use LiteTrack-B6 \cite{litetrack} and LiteTrack-B8 \cite{litetrack} as pre-trained models for STFTrack-S and STFTrack-B, respectively.

\textbf{Second Stage Training.} We train the MTFM module. The training set contains LasHeR \cite{lasher}, where RGB images are converted to grayscale images, and thermal images are used to simulate acoustic binary images. The training epoch number is 15, which takes about 2 hours. The total learning rate is $2 \times 10^{-5}$.

\subsection{Comparison Results}
\subsubsection{General Trackers}
We evaluate general baseline trackers on the SonarT165 benchmark. The results are reported in Table \ref{table baseline results}. The PR of general trackers is mostly around 80\%, which means that the simple appearance of acoustic targets does not pose a significant challenge to current trackers. However, the best SR score among these trackers is 57.4\% (achieved by ARTrackV2Seq-B256 \cite{artrackv2}), which means that strong background noise and weaker texture information in acoustic images are challenging for current trackers. Similarly, OP50 and OP75 also reflect this issue, especially since most current trackers have an OP75 score below 30\%. This means that the trackers still have great potential for improvement in achieving precise acoustic object tracking.

We compare the performance of STFTrack-B and general trackers. In fan sequences, STFTrack-B outperforms ARTrackV2Seq-B256 \cite{artrackv2} and LoRAT-L224 \cite{lorat} by 2.0\% in SR, 2.9\% and 1.8\% in PR, 1.5\% and 0.7\% in NPR, respectively. In square sequences, STFTrack-B outperforms ARTrackv2Seq-B256 \cite{artrackv2} and ARTrackSeq-B256 \cite{artrack} by 1.7\% and 2.3\% in SR, both 2.5\% in PR, respectively. In addition, STFTrack's SR, OP50, OP75, NPR, and F1 scores in fan sequences are better than square sequences, which means that it is more suitable for acoustic object tracking in fan sequences. Overall, STFTrack-B achieves state-of-the-art performance among general trackers.

\input{tables/baselines_lightweight_results}

\subsubsection{Lightweight Trackers}
We evaluate lightweight baseline trackers on the SonarT165 benchmark. The results are reported in Table \ref{table baseline lightweigh tresults}. The SR and PR scores of the state-of-the-art lightweight trackers (such as SMAT \cite{smat}, LiteTrack \cite{litetrack}) are not significantly lower than those of the advanced general trackers, which means that the development of acoustic trackers based on lightweight trackers is more suitable for the UAOT task.

Also, we compare the performance of STFTrack-S and lightweight trackers. In fan sequences, STFTrack-S outperforms LiteTrack-B8 \cite{litetrack} and LiteTrack-B9 \cite{litetrack} by 3.8\% and 4.2\% in SR, 6.2\% and 5.4\% in PR, 5\% and 2.6\% in NPR, respectively. In square sequences, STFTrack-S outperforms LiteTrack-B8 \cite{litetrack} and LiteTrack-B9 \cite{litetrack} by 1.5\% and 2.4\% in SR, 4.5\% and 6.0\% in PR, 3.8\% and 2.5\% in NPR, respectively. Similar to STFTrack-B, it also has better performance in fan sequences. Overall, STFTrack-S achieves state-of-the-art performance among lightweight trackers.

\subsection{Attribute Studies}
We present the attribute results of STFTrack-B, STFTrack-S, and their baselines LiteTrack-B8 \cite{litetrack} and LiteTrack-B6 \cite{litetrack} in Figure \ref{fig: radamap}. In terms of SR, our method demonstrates better scores in scale variation (SV), field environment (FE) attributes, while further improvement is needed in acoustic object crossover (AOC), small target (ST), out-of-view (OV) and low acoustic reflection (LAR) attributes. Similar problems also exist in terms of PR and NPR. In addition, compared to the baseline method \cite{litetrack}, STFTrack achieves significant performance improvements on each attribute.

\input{tables/ablation_dataset}
\input{tables/ablation_freq_enhance}
\input{tables/ablation_template_fusion}
\input{tables/ablation_tarj_fusion}
\input{tables/ablation_sonar_img}
\subsection{Ablation Studies}
We explore the effectiveness of STFTrack-B components through ablation experiments on the SonarT165 benchmark. In the ablation experiments, we report the SR score, the PR score, and the NPR score of the tracker in two types of sequence.

\subsubsection{Ablation of Training Datasets}
We evaluate the contributions of different training datasets, as shown in Table \ref{table ablation datasets}. The use of LaSOT \cite{lasot}, GOT10K \cite{got10k}, and UATD \cite{uatd} training sets effectively improved the model's adaptability to acoustic images. In addition, the widely used COCO \cite{coco}, TrackingNet \cite{trackingnet}, and SARDet \cite{sardet} training sets are unable to produce gains in the model. Overall, the three datasets used have a positive impact on the model.

\subsubsection{Ablation of FEM Module}
We evaluate the contributions of each component of the FEM module, as shown in Table \ref{fig: freq enhance module}. Using only high-frequency feature enhancement effectively improves the model performance; however, using only low-frequency features reduces the discriminative ability of the model. Tracking performance is further improved when high-frequency enhancement is combined with low-frequency enhancement. Overall, they all play an important role.

\subsubsection{Ablation of MTFM Module}
We evaluate the contributions of each component of the MTFM module, as shown in Table \ref{table ablation template fuision}. The integration of dual templates improves the performance of the model in both square and fan sequences. In addition, multi-view integration of the dynamic template also plays an important role, bringing a score improvement of 1.1\% SR and 0.7\% SR to square and fan sequences, respectively. Overall, each component of the MTFM module contributes to improving performance.

\subsubsection{Ablation of OTCM Module}
We evaluate the contributions of each component of the OTCM module, as shown in Table \ref{table ablation traj fusion}. 
The introduction of the IoU score $I_{box}$ and the brightness response $R_{np}$ improved the performance of the model, respectively. In addition, the IoB score $I_{M}$ provides a slight performance boost at a negligible computational cost. Overall, each component of the OTCM module contributes to improving performance.

\subsubsection{Ablation of Image Enhancement}
We evaluate the contributions of each component and the different variants of the acoustic image enhancement module, as shown in Table \ref{table ablation sonar img}.
Two high-frequency enhancements to the acoustic image effectively improve the tracker's performance in both the fan and square sequences. However, this operation is not optimal in the fan sequence, and three high-frequency enhancements achieve higher scores, but it reduces the performance of the model in the square sequence. Overall, the introduction of acoustic image enhancement methods plays an important role in improving acoustic tracker performance. We hope that this work can promote researchers' attention to adaptive enhancement methods for acoustic images.

\input{figs/vis_tracking_results}
\input{figs/vis_heatmap}
\input{figs/vis_failure_cases}
\subsection{Visualization}
\subsubsection{Heatmap}
We present some heat maps of STFTrack and its baseline \cite{litetrack}, as shown in Figure \ref{fig: heatmap}. In general sequences (Figure \ref{fig: heatmap} (1) and (4)), STFTrack exhibits better feature attention than LiteTrack \cite{litetrack}. When the target has a low acoustic reflection value (Figure \ref{fig: heatmap} (2)), the feature focus of the LiteTrack \cite{litetrack} model diverges, while our method can still maintain the discrimination of the target's appearance. When there are sound crossing objects around the target (Figure \ref{fig: heatmap} (3)), our method demonstrates better robustness and discrimination than the baseline \cite{litetrack}. Overall, STFTrack demonstrates significant improvements in feature attention.

\subsubsection{Tracking Results}
We present some tracking results for STFTrack-B, LiteTrack-B8 \cite{litetrack}, ARTrackV2Seq-B256 \cite{artrackv2}, and LoRAT-L224 \cite{lorat} on six representative sequences in SonarT165 benchmark, as shown in Figure \ref{fig: trackingresults}. When the target reappears after out of view (Figure \ref{fig: trackingresults} (a)), LiteTrack-B8 \cite{litetrack} and LoRAT-L224 \cite{lorat} lose the target, while our method still accurately tracks the target. When the target is affected by background interference (Figure \ref{fig: trackingresults} (b)), our method still maintains accuracy, while other trackers track drift. The acoustic ghosting of the target causes other trackers to produce inaccurate bounding boxes (Figure \ref{fig: trackingresults} (c)), while STFTrack-B shows stronger robustness. Similarly, when there is interference from similar objects around  (Figure \ref{fig: trackingresults} (d)), our method can still accurately track, while other trackers drift or experience a decrease in accuracy. Finally, compared to other methods, STFTrack-B demonstrates better adaptability in outdoor environments (Figure \ref{fig: trackingresults} (e-f)). Overall, STFTrack achieves better acoustic object tracking.

\subsubsection{Failure Cases}
We present some typical failure cases of STFTrack-B on six representative sequences in SonarT165 benchmark, as shown in Figure \ref{fig: fail cases}. STFTrack is prone to accuracy degradation (Figure \ref{fig: fail cases} (a)(b)(f)) or tracking drift (Figure \ref{fig: fail cases} (e)) when subjected to background interference. The long-term out-of-view of the target is also challenging for our tracker Figure \ref{fig: fail cases} (c). In addition, the intersection of acoustic objects can also cause tracking drift in our tracker (Figure \ref{fig: fail cases} (d)).

%% file: tables/baselines_results.tex
\begin{table*}[]
\caption{Comparison results for our method and general trackers in the proposed benchmark. The best three results are shown in \color{red}{red}\color{black}{,} \color{blue}{blue} \color{black}and \color{green}green \color{black} fonts.}
\label{table baseline results}
\centering
\renewcommand{\arraystretch}{1.1}
\setlength{\tabcolsep}{2.5pt}
\begin{tabular}{l|c|cccccc|cccccc|cccccc}
\hline
\multirow{2}{*}{General Tracker} & \multirow{2}{*}{Year} & \multicolumn{6}{c|}{SonarT165} & \multicolumn{6}{c|}{SonarT165-Fan} & \multicolumn{6}{c}{SonarT165-Square} \\ \cline{3-20} 
                                    &      & SR   & OP50 & OP75 & PR   & NPR  & F1   & SR   & OP50 & OP75 & PR   & NPR  & F1   & SR   & OP50 & OP75 & PR   & NPR  & F1   \\ \hline
SiamRPN \cite{siamrpn}              & 2018 & 48.0 & 60.3 & 14.5 & 78.9 & 58.5 & 72.0 & 48.8 & 62.8 & 17.3 & 76.8 & 59.8 & 74.3 & 47.2 & 57.8 & 11.7 & 81.1 & 57.1 & 69.6 \\ \hline
SiamRPN++ \cite{siamrpn++}          & 2019 & 53.0 & 66.8 & 16.6 & 86.9 & 62.7 & 76.9 & 53.6 & 68.3 & 18.7 & 85.5 & 64.6 & 76.4 & 52.3 & 65.3 & 14.6 & \color{blue}\textbf{88.3} & 60.9 & 77.3 \\
DiMP18 \cite{dimp}                  & 2019 & 50.6 & 61.6 & 19.3 & 84.1 & \color{blue}\textbf{71.1} & 73.1 & 49.5 & 60.3 & 20.1 & 80.7 & 69.2 & 71.8 & 51.6 & 62.9 & 18.5 & 87.4 & 73.0 & 74.5 \\
DiMP50 \cite{dimp}                  & 2019 & 53.1 & 67.4 & 21.5 & 83.8 & 69.4 & 78.9 & 51.0 & 64.5 & 22.0 & 79.3 & 66.4 & 76.4 & 55.1 & 70.3 & 21.0 & \color{blue}\textbf{88.3} & 72.5 & 81.4 \\ \hline
PrDiMP18 \cite{prdimp}              & 2020 & 45.0 & 54.0 & 19.6 & 76.5 & 61.5 & 65.0 & 41.8 & 48.3 & 17.3 & 73.7 & 59.0 & 60.0 & 48.3 & 59.8 & 21.9 & 79.3 & 64.0 & 69.6 \\
PrDiMP50 \cite{prdimp}              & 2020 & 47.6 & 59.5 & 21.3 & 74.7 & 60.7 & 70.6 & 45.1 & 56.0 & 19.7 & 71.2 & 58.9 & 67.3 & 50.1 & 63.1 & 22.9 & 78.3 & 62.6 & 73.7 \\
SiamCAR \cite{siamcar}              & 2020 & 47.7 & 59.2 & 13.3 & 80.0 & 53.0 & 70.0 & 46.9 & 58.4 & 13.6 & 77.4 & 52.6 & 69.0 & 48.4 & 59.9 & 12.9 & 82.5 & 53.4 & 70.9 \\
SiamBAN \cite{siamban}              & 2020 & 53.2 & 67.5 & 21.0 & 84.2 & 64.0 & 78.6 & 53.3 & 68.0 & 22.1 & 83.1 & 65.3 & 79.6 & 53.1 & 67.1 & 19.9 & 85.4 & 62.6 & 77.7 \\
SiamBAN-ACM \cite{siambanacm}       & 2020 & 53.2 & 66.6 & 21.6 & 84.4 & 61.8 & 77.8 & 53.1 & 67.1 & 21.1 & 82.9 & 62.9 & 77.4 & 53.3 & 66.2 & 22.0 & 85.8 & 60.6 & 78.2 \\ \hline
KeepTrack \cite{keeptrack}          & 2021 & 47.9 & 59.6 & 19.6 & 76.8 & 58.2 & 73.0 & 46.5 & 57.6 & 18.6 & 74.0 & 57.2 & 71.5 & 49.2 & 61.5 & 20.5 & 79.6 & 59.2 & 74.5 \\
TrDiMP50 \cite{trdimp}              & 2021 & 49.7 & 61.7 & 20.2 & 80.0 & 62.0 & 74.5 & 47.4 & 58.5 & 18.7 & 76.8 & 61.1 & 72.4 & 52.0 & 64.9 & 21.7 & 83.1 & 63.0 & 76.5 \\
TransT \cite{transt}                & 2021 & 49.3 & 61.6 & 18.2 & 80.2 & 63.6 & 72.1 & 51.4 & 65.0 & 19.9 & 81.2 & 65.2 & 74.0 & 47.3 & 58.3 & 16.1 & 79.2 & 62.1 & 70.2 \\
StarkS50 \cite{stark}               & 2021 & 43.0 & 54.5 & 16.4 & 68.0 & 49.9 & 66.0 & 42.9 & 54.7 & 18.3 & 65.1 & 49.8 & 65.9 & 43.2 & 54.3 & 14.6 & 70.9 & 49.9 & 66.1 \\
StarkST50 \cite{stark}              & 2021 & 46.4 & 58.7 & 17.8 & 73.5 & 54.5 & 69.3 & 48.4 & 61.9 & 21.6 & 73.3 & 56.7 & 73.3 & 44.4 & 55.5 & 14.0 & 73.8 & 52.3 & 65.1 \\
StarkST101 \cite{stark}             & 2021 & 45.4 & 56.2 & 16.6 & 73.8 & 53.4 & 68.3 & 45.9 & 57.9 & 19.3 & 71.3 & 53.1 & 69.4 & 44.8 & 54.5 & 13.9 & 76.3 & 53.6 & 67.1 \\ \hline
ToMP50 \cite{tomp}                  & 2022 & 53.2 & 68.1 & \color{red}\textbf{27.8} & 79.4 & 65.1 & 77.3 & 52.1 & 67.1 & 26.8 & 77.6 & 64.0 & 76.3 & 54.3 & 69.2 & \color{red}\textbf{28.8} & 81.2 & 66.2 & 78.3 \\
ToMP101 \cite{tomp}                 & 2022 & 52.6 & 67.0 & 26.4 & 79.4 & 64.6 & 76.8 & 53.1 & 68.1 & 26.0 & 79.7 & 65.7 & 77.5 & 52.1 & 65.9 & \color{blue}\textbf{26.9} & 79.1 & 63.6 & 76.1 \\
OSTrack256 \cite{ostrack}           & 2022 & 54.0 & 69.2 & 22.1 & 84.3 & 65.8 & 79.4 & 55.4 & 72.3 & 25.9 & 83.0 & 67.7 & 80.7 & 52.7 & 66.1 & 18.4 & 85.6 & 63.9 & 78.1 \\
OSTrack386 \cite{ostrack}           & 2022 & 49.2 & 63.1 & 19.4 & 77.4 & 61.8 & 75.5 & 48.7 & 63.1 & 22.2 & 74.1 & 61.4 & 74.7 & 49.8 & 63.0 & 16.5 & 80.7 & 62.2 & 76.4 \\
AiATrack \cite{aiatrack}            & 2022 & 48.0 & 60.2 & 18.9 & 77.1 & 53.0 & 71.6 & 48.4 & 61.0 & 19.4 & 76.0 & 53.6 & 72.0 & 47.6 & 59.4 & 18.4 & 78.1 & 52.3 & 71.2 \\ \hline
UOSTrack \cite{uostrack}            & 2023 & 54.6 & 69.4 & 21.6 & 86.3 & 65.6 & 80.4 & 55.8 & 72.0 & 26.0 & 84.5 & 66.9 & 81.3 & 53.5 & 66.7 & 17.1 & \color{green}\textbf{88.1} & 64.3 & 79.4 \\ 
ARTrackSeq-B256 \cite{seqtrack}     & 2023 & 55.5 & 71.1 & 22.9 & 86.3 & 67.9 & 81.8 & 55.2 & 70.5 & 24.5 & 84.4 & 67.5 & 80.1 & \color{green}\textbf{55.8} & 71.8 & 21.2 & \color{blue}\textbf{88.3} & 68.3 & \color{blue}\textbf{83.4} \\
SeqTrack-B256 \cite{seqtrack}       & 2023 & 46.0 & 57.9 & 15.6 & 74.3 & 57.7 & 70.6 & 45.7 & 58.8 & 17.5 & 70.9 & 56.5 & 69.7 & 46.4 & 57.0 & 13.8 & 77.7 & 59.0 & 71.4 \\
SeqTrack-B384 \cite{seqtrack}       & 2023 & 46.1 & 57.3 & 15.9 & 76.6 & 59.7 & 70.5 & 45.6 & 57.7 & 18.0 & 73.0 & 58.7 & 69.8 & 46.6 & 56.9 & 13.8 & 80.1 & 60.7 & 71.3 \\
SeqTrack-L256 \cite{seqtrack}       & 2023 & 46.3 & 58.7 & 17.2 & 73.9 & 57.5 & 71.3 & 45.7 & 58.4 & 19.1 & 70.7 & 56.8 & 70.4 & 47.0 & 59.1 & 15.4 & 77.1 & 58.1 & 72.2 \\
SeqTrack-L384 \cite{seqtrack}       & 2023 & 47.1 & 59.6 & 16.3 & 76.3 & 59.6 & 72.1 & 46.9 & 59.5 & 17.8 & 74.2 & 59.5 & 71.4 & 47.3 & 59.6 & 14.8 & 78.4 & 59.6 & 72.8 \\
HiPTrack \cite{hiptrack}            & 2023 & 55.1 & 71.5 & 24.6 & 84.5 & 65.5 & 80.7 & 55.3 & 71.6 & 26.5 & 82.9 & 65.4 & 80.6 & 54.9 & 71.4 & 22.7 & 86.0 & 65.6 & 80.9 \\ \hline
ODTrack-B256 \cite{odtrack}         & 2024 & 54.6 & 71.5 & 21.2 & 85.9 & 70.2 & 81.4 & 54.7 & 71.1 & 22.1 & 85.3 & \color{green}\textbf{72.3} & 80.9 & 54.6 & \color{green}\textbf{71.9} & 20.3 & 86.6 & 68.1 & \color{green}\textbf{81.5} \\
ODTrack-L256 \cite{odtrack}         & 2024 & 53.1 & 69.2 & 17.9 & 84.7 & 69.4 & 81.0 & 52.7 & 68.7 & 20.5 & 82.4 & 70.3 & 80.3 & 53.6 & 69.6 & 17.6 & 87.1 & 68.5 & 81.4 \\
ARTrackv2Seq-B256 \cite{seqtrack}   & 2024 & \color{blue}\textbf{57.4} & \color{blue}\textbf{74.0} & 25.5 & \color{blue}\textbf{88.1} & 70.5 & \color{red}\textbf{84.4} & \color{blue}\textbf{58.3} & 75.3 & \color{green}\textbf{27.7} & \color{green}\textbf{88.0} & 72.1 & \color{red}\textbf{84.2} & \color{blue}\textbf{56.4} & \color{blue}\textbf{72.7} & 23.2 & \color{blue}\textbf{88.3}& \color{green}\textbf{68.9} & \color{red}\textbf{84.5} \\
LoRAT-B224 \cite{lorat}             & 2024 & 52.3 & 67.7 & 19.4 & 82.7 & 64.3 & 77.7 & 52.5 & 67.7 & 22.0 & 81.3 & 63.6 & 78.3 & 52.1 & 67.7 & 16.8 & 84.2 & 64.9 & 77.2 \\
LoRAT-B378 \cite{lorat}             & 2024 & 51.5 & 66.2 & 19.3 & 81.9 & 64.5 & 75.9 & 51.6 & 66.4 & 20.7 & 80.9 & 63.7 & 75.1 & 51.4 & 65.9 & 17.9 & 82.9 & 65.4 & 76.7 \\
LoRAT-L224 \cite{lorat}             & 2024 & \color{green}\textbf{56.2} & \color{green}\textbf{73.7} & 22.9 & \color{green}\textbf{87.2} & 70.3 & \color{green}\textbf{82.2} & \color{blue}\textbf{58.3} & \color{blue}\textbf{76.5} & 26.1 & \color{blue}\textbf{89.1} & \color{blue}\textbf{72.9} & \color{red}\textbf{84.2} & 54.1 & 71.0 & 19.8 & 85.4 & 67.8 & 80.1 \\
LoRAT-L378 \cite{lorat}             & 2024 & 55.2 & 72.2 & 22.3 & 86.5 & \color{green}\textbf{70.9} & 80.5 & 56.0 & 73.3 & 24.1 & 86.5 & 71.9 & 80.9 & 54.5 & 71.1 & 20.4 & 86.6 & \color{red}\textbf{69.9} & 80.2 \\
LoRAT-G224 \cite{lorat}             & 2024 & 54.9 & 72.0 & 23.9 & 84.3 & 68.5 & 80.5 & \color{green}\textbf{57.6} & \color{green}\textbf{75.4} & \color{blue}\textbf{27.9} & 86.5 & 70.9 & \color{green}\textbf{83.1} & 52.3 & 68.7 & 19.9 & 82.0 & 66.1 & 77.8 \\
LoRAT-G378 \cite{lorat}             & 2024 & 54.8 & 71.6 & 23.0 & 84.4 & 67.8 & 80.2 & 55.4 & 72.1 & 25.3 & 84.0 & 68.0 & 80.2 & 54.2 & 71.1 & 20.6 & 84.7 & 67.5 & 80.1 \\
MCITrack-B224 \cite{mcitrack}       & 2024 & 49.0 & 62.2 & 24.6 & 74.0 & 58.6 & 72.2 & 48.6 & 62.2 & 25.5 & 71.9 & 58.5 & 70.7 & 49.5 & 62.2 & 23.7 & 76.1 & 58.6 & 73.6 \\
MCITrack-L224 \cite{mcitrack}       & 2024 & 49.2 & 62.7 & 24.0 & 75.0 & 59.5 & 72.8 & 48.0 & 61.5 & 24.9 & 71.4 & 58.3 & 69.7 & 50.4 & 63.9 & 23.0 & 78.7 & 60.7 & 75.8 \\
MCITrack-L384 \cite{mcitrack}       & 2024 & 51.7 & 65.8 & \color{green}\textbf{25.6} & 78.8 & 62.8 & 76.2 & 51.3 & 65.5 & 27.3 & 76.6 & 62.0 & 75.3 & 52.1 & 66.0 & \color{green}\textbf{23.8} & 81.0 & 63.7 & 77.2 \\ \hline
STFTrack-B256                       & -    & \color{red}\textbf{59.2} & \color{red}\textbf{76.4} & \color{blue}\textbf{26.7} & \color{red}\textbf{90.8} & \color{red}\textbf{71.3} & \color{blue}\textbf{82.8} & \color{red}\textbf{60.3} & \color{red}\textbf{77.8} & \color{red}\textbf{29.7} & \color{red}\textbf{90.9} & \color{red}\textbf{73.6} & \color{blue}\textbf{84.0} & \color{red}\textbf{58.1} & \color{red}\textbf{75.1} & \color{green}\textbf{23.8} & \color{red}\textbf{90.8} & \color{blue}\textbf{69.0} & \color{green}\textbf{81.5} \\ \hline
\end{tabular}
\end{table*}

%% file: figs/exp_plots.tex
\begin{figure*}[] %
\begin{minipage}[]{0.33\linewidth} 
\centering
\includegraphics[width=6cm,height=4.5cm]{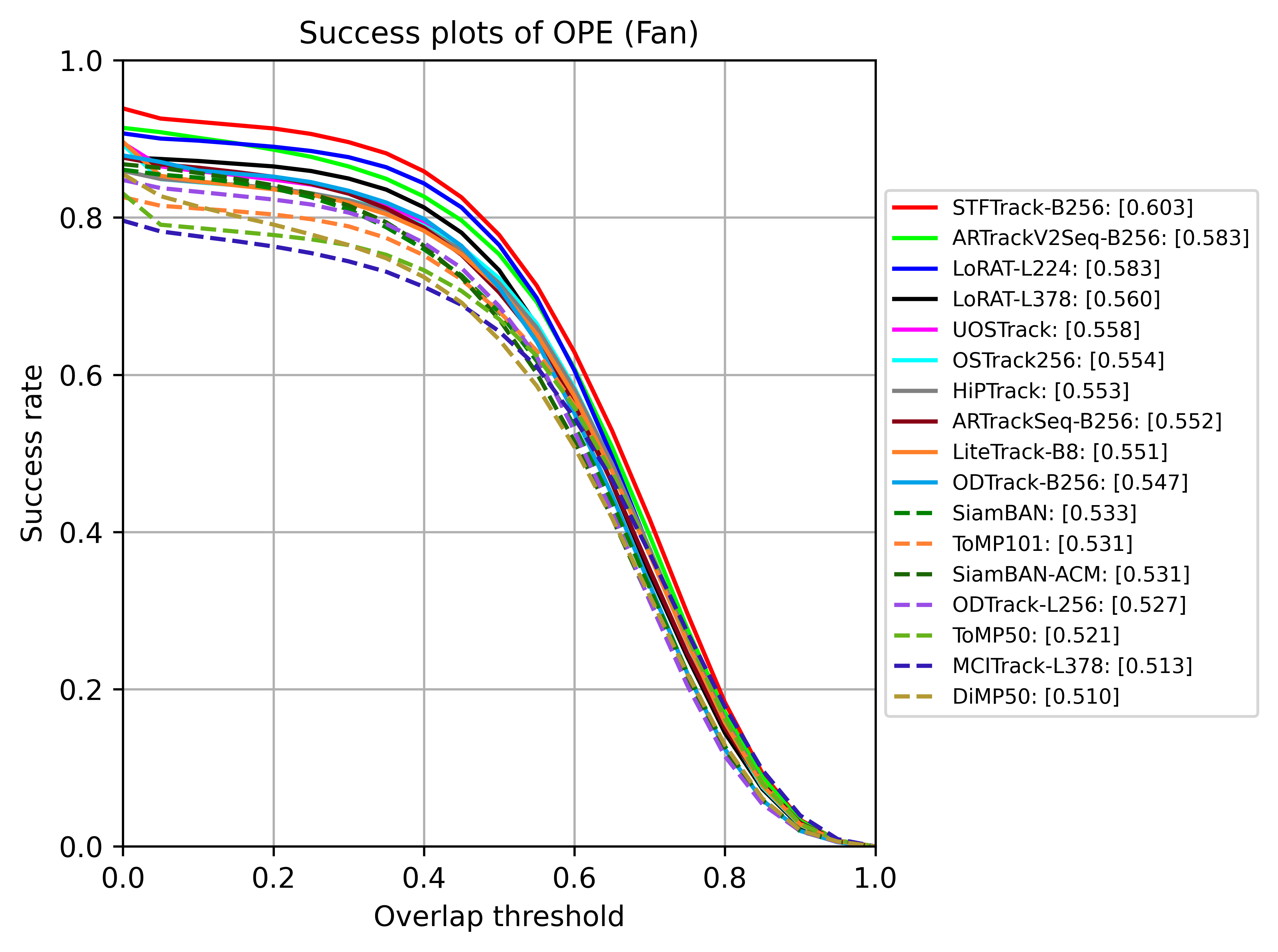} 
\end{minipage}%
\begin{minipage}[]{0.33\linewidth}
\centering
\includegraphics[width=6cm,height=4.5cm]{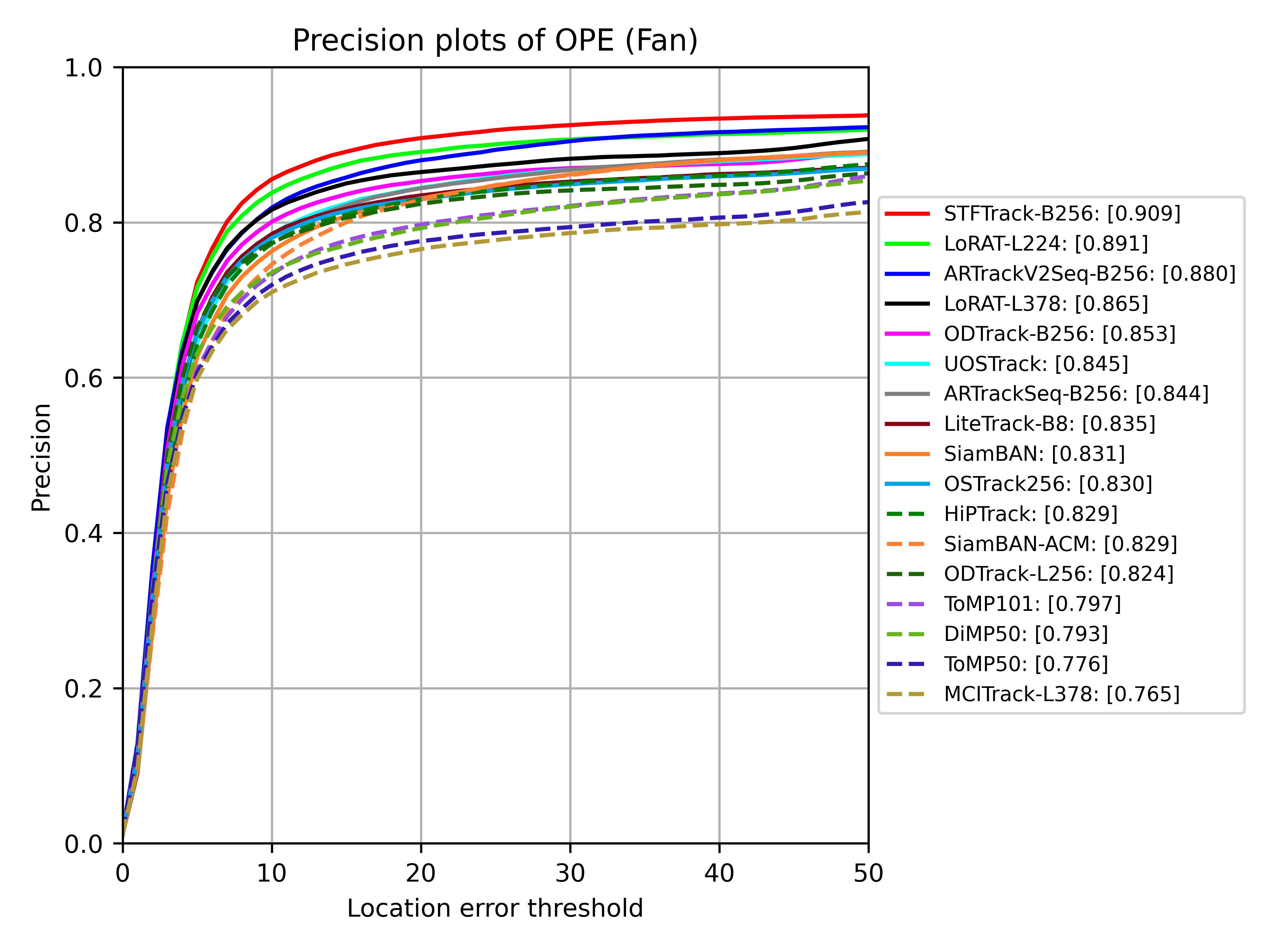}
\end{minipage}%
\begin{minipage}[]{0.33\linewidth}
\centering
\includegraphics[width=6cm,height=4.5cm]{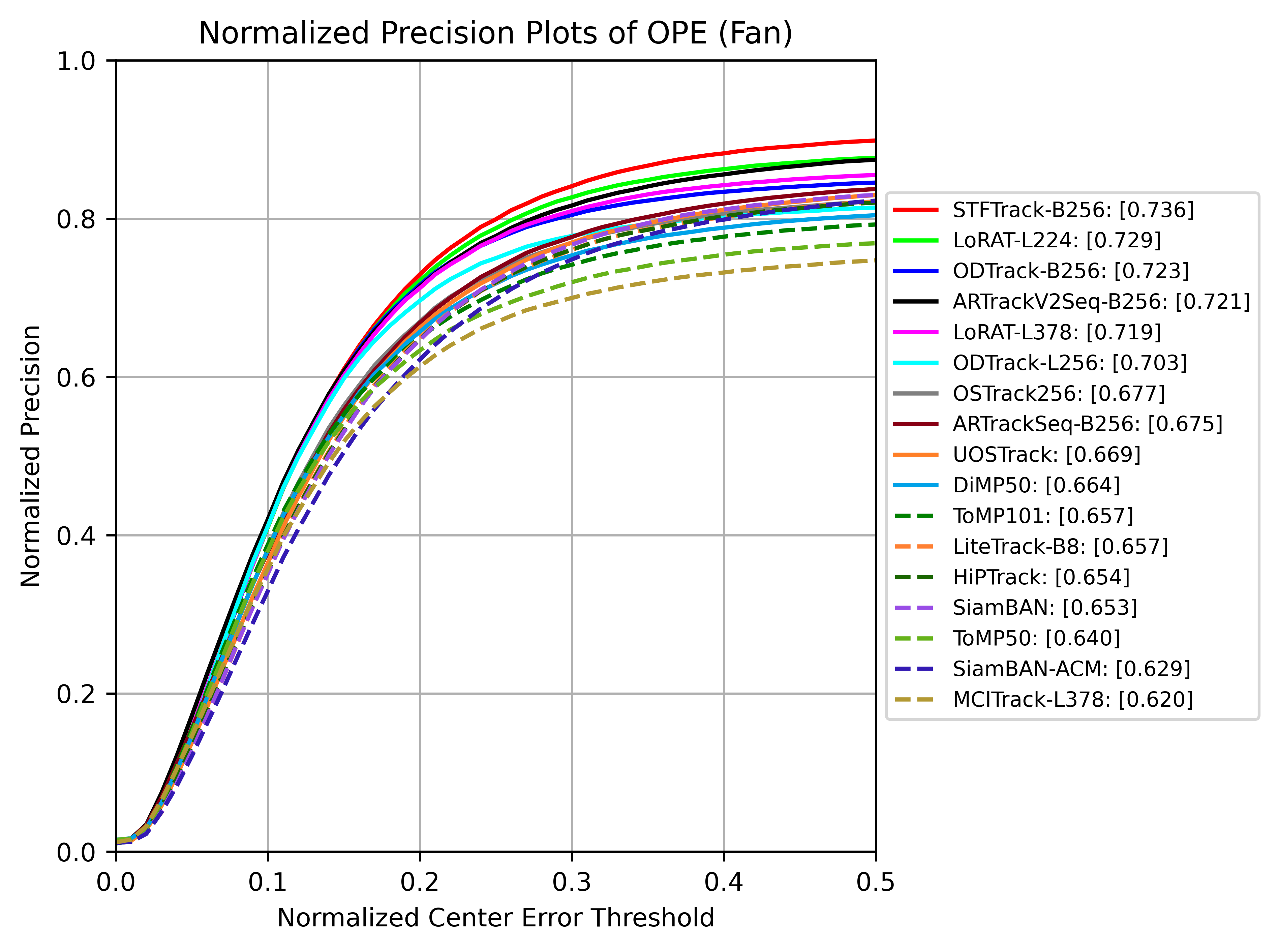}
\end{minipage}
\begin{minipage}[]{0.33\linewidth} 
\centering
\includegraphics[width=6cm,height=4.5cm]{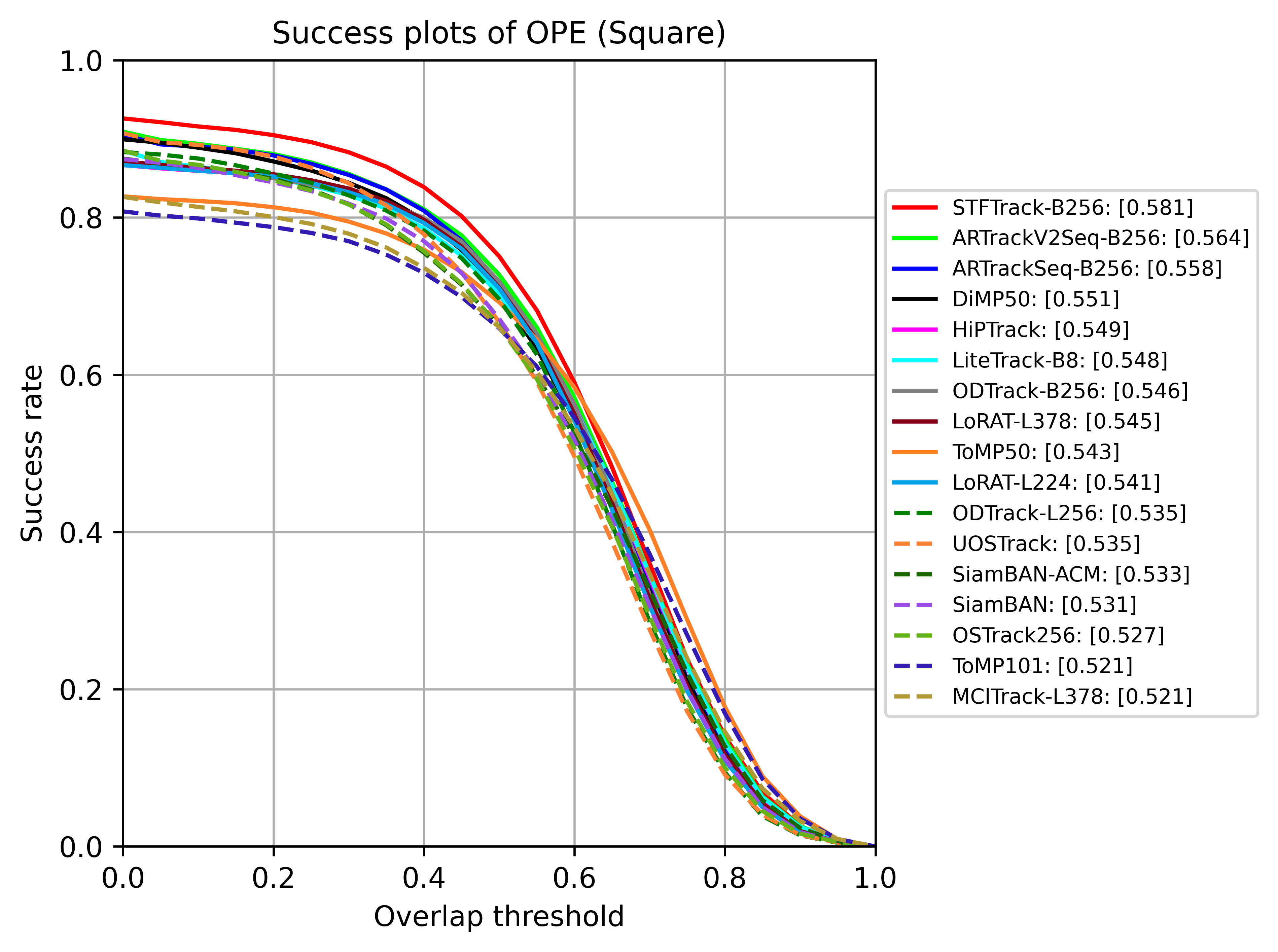} 
\end{minipage}%
\begin{minipage}[]{0.33\linewidth}
\centering
\includegraphics[width=6cm,height=4.5cm]{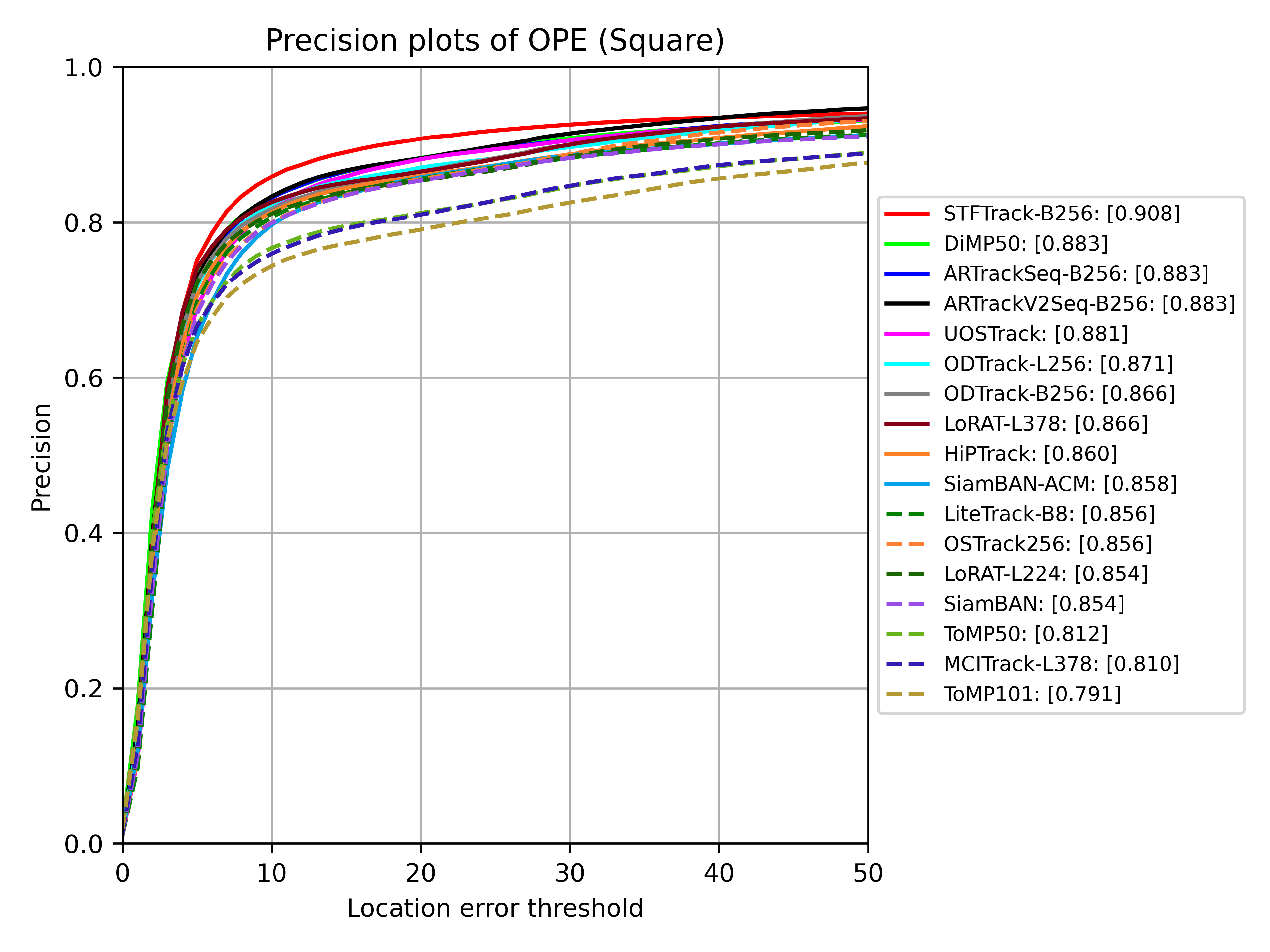}
\end{minipage}%
\begin{minipage}[]{0.33\linewidth}
\centering
\includegraphics[width=6cm,height=4.5cm]{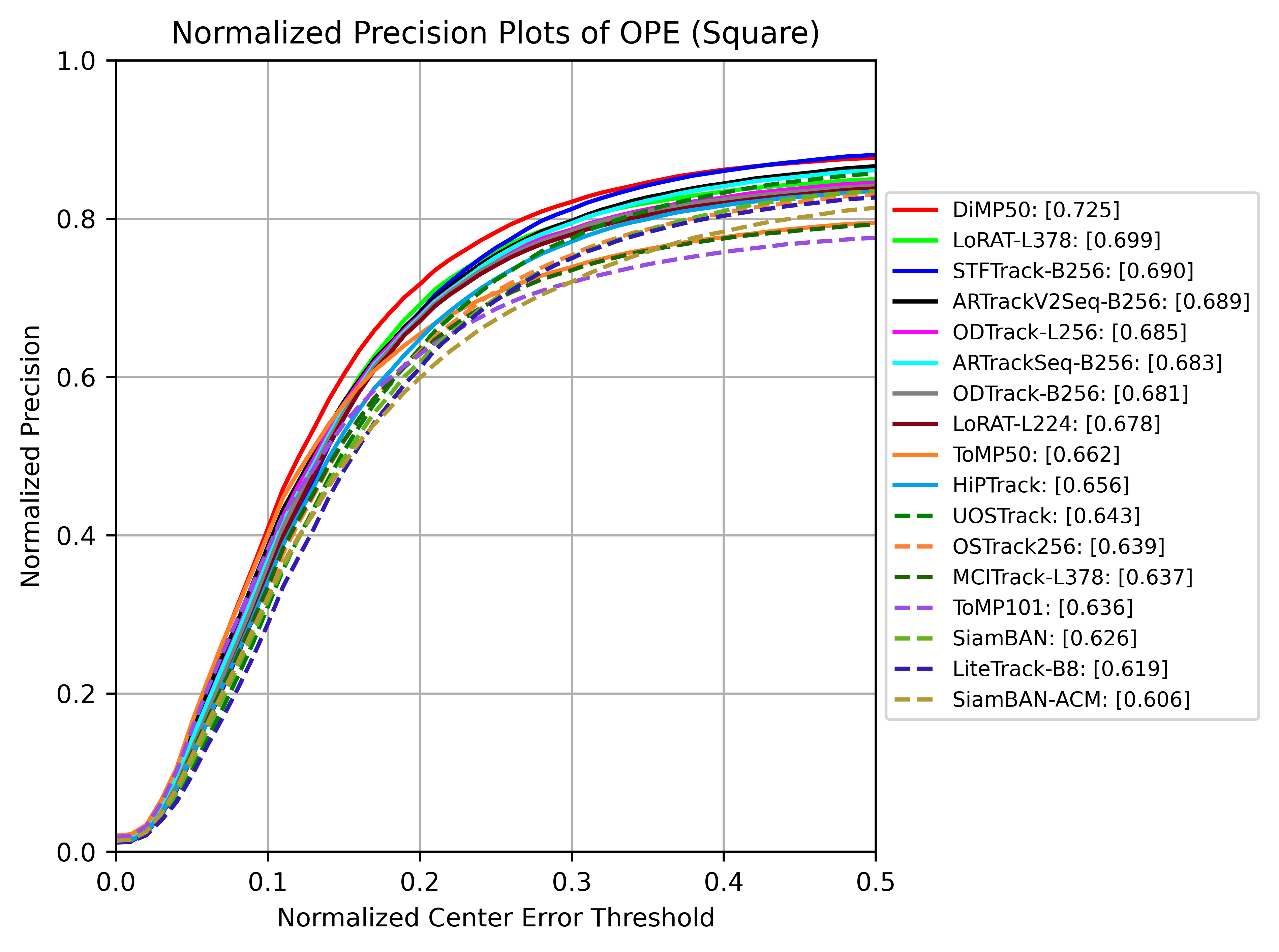}
\end{minipage}
\caption{The success plots, precision plots and normalized precision plots of the trackers. These trackers are STFTrack-B256, ARTrackV2Seq-B256 \cite{artrackv2}, LoRAT-L224 \cite{lorat}, LoRAT-L378 \cite{lorat}, UOSTrack \cite{uostrack}, OSTrack256 \cite{ostrack}, HiPTrack \cite{hiptrack}, ARTrackSeq-B256 \cite{artrack}, LiteTrack \cite{litetrack}, ODTrack-B256 \cite{odtrack}, ODTrack-L256 \cite{odtrack}, SiamBAN \cite{siamban}, ToMP50 \cite{tomp}, ToMP101 \cite{tomp}, SiamBAN-ACM \cite{siambanacm}, MCITrack-L378 \cite{mcitrack}, DiMP50 \cite{dimp}.}
\label{fig: rgbs_plots}

\end{figure*}

%% file: figs/exp_attr_radamaps.tex
\begin{figure*}[] %
\begin{minipage}[]{0.33\linewidth} 
\centering
\includegraphics[width=5.5cm,height=5.5cm]{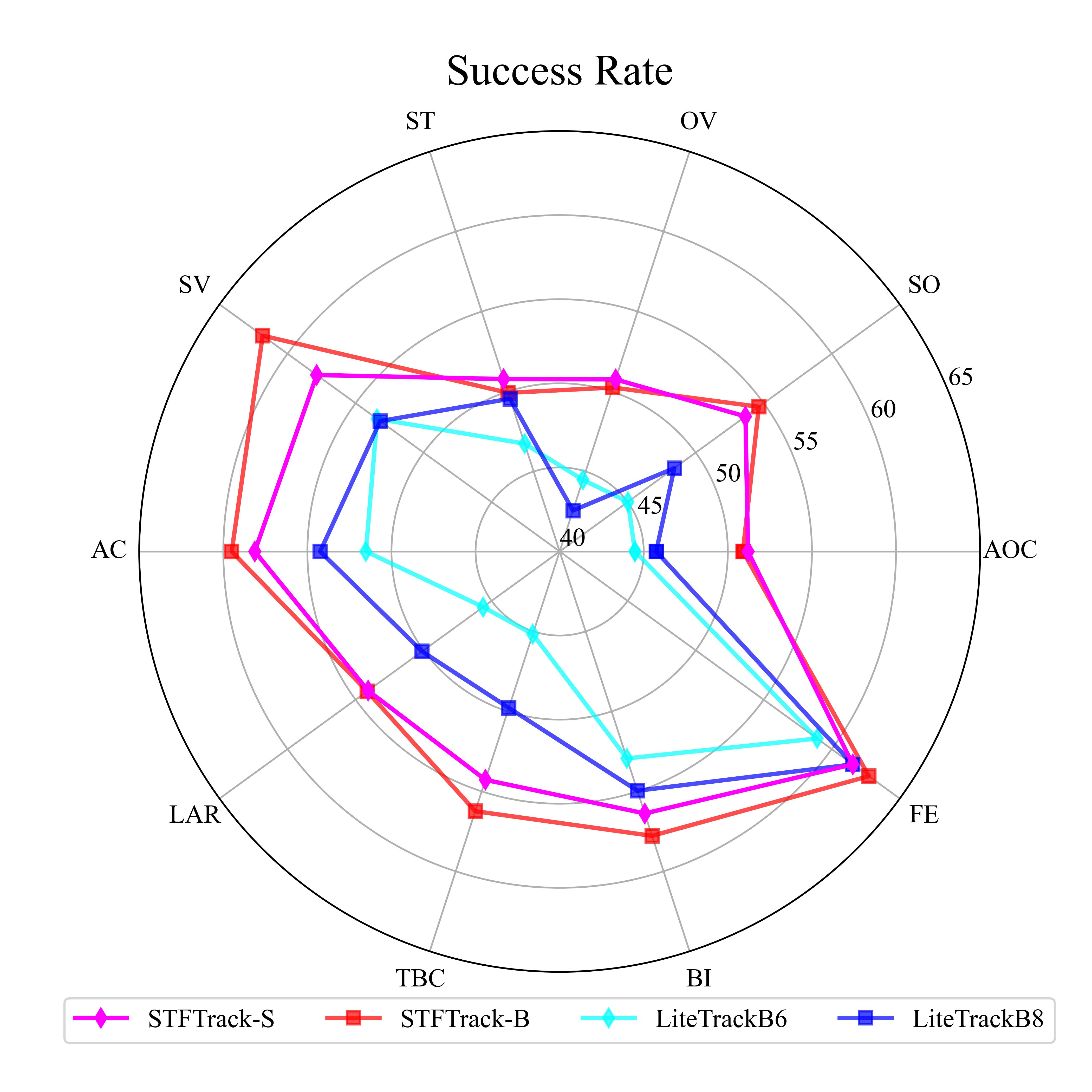} 
\end{minipage}%
\begin{minipage}[]{0.33\linewidth}
\centering
\includegraphics[width=5.5cm,height=5.5cm]{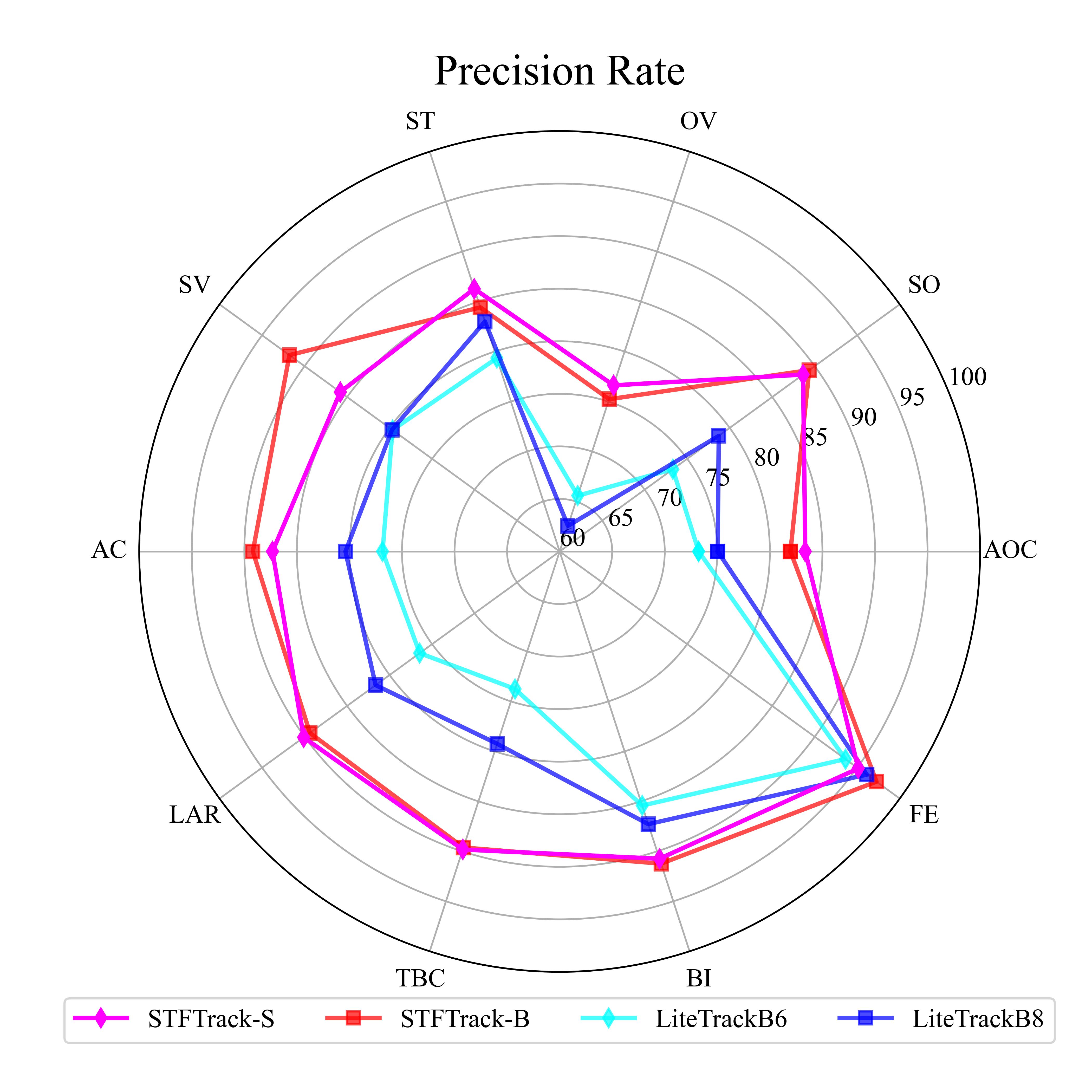}
\end{minipage}%
\begin{minipage}[]{0.33\linewidth}
\centering
\includegraphics[width=5.5cm,height=5.5cm]{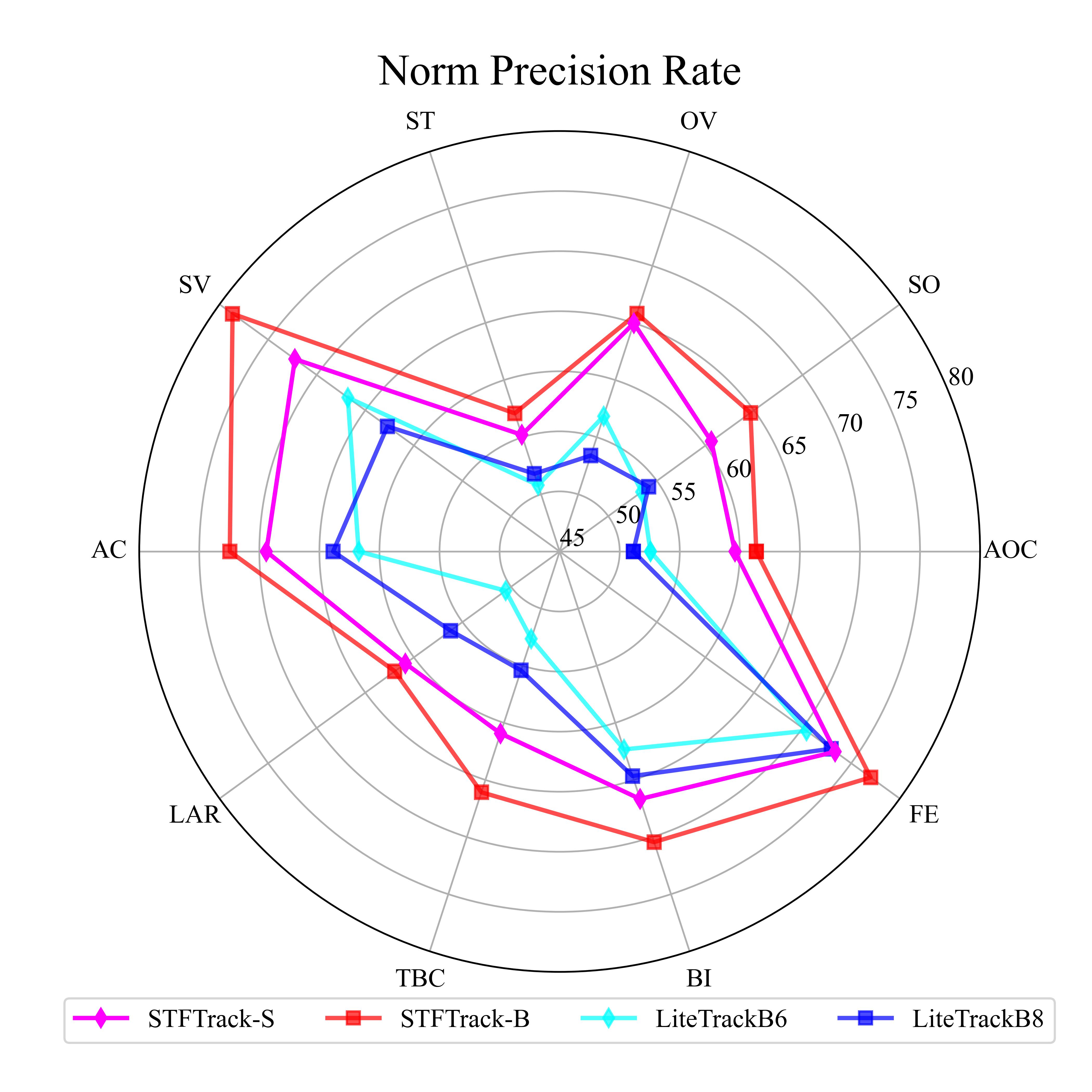}
\end{minipage}

\caption{The success rate, precision rate and normalized precision rate of STFTrack-B, STFTrack-S, LiteTrack-B8 \cite{litetrack}, LiteTrack-B6 \cite{litetrack} under different attributes on SonarT165 benchmark.}
\label{fig: radamap}

\end{figure*}

%% file: tables/baselines_lightweight_results.tex
\begin{table*}[]
\caption{Comparison results for our method and lightweight trackers in the proposed benchmark. The best three results are shown in \color{red}{red}\color{black}{,} \color{blue}{blue} \color{black}and \color{green}green \color{black} fonts.}
\label{table baseline lightweigh tresults}
\centering
\renewcommand{\arraystretch}{1.1}
\setlength{\tabcolsep}{2.5pt}
\begin{tabular}{l|c|cccccc|cccccc|cccccc}
\hline
\multirow{2}{*}{Lightweight Tracker} & \multirow{2}{*}{Year} & \multicolumn{6}{c|}{SonarT165} & \multicolumn{6}{c|}{SonarT165-Fan} & \multicolumn{6}{c}{SonarT165-Square} \\ \cline{3-20} 
     &                              & AUC  & OP50 & OP75 & PR   & NPR  & F1   & AUC  & OP50 & OP75 & PR   & NPR  & F1   & AUC  & OP50 & OP75  & PR   & NPR  & F1   \\ \hline
MobileSiamRPN++ \cite{siamrpn++}    & 2019  & 48.6 & 62.2 & 14.8 & 79.5 & 58.6 & 73.4 & 48.9 & 62.7 & 17.4 & 77.5 & 59.1 & 74.1 & 48.3 & 61.7 & 12.2  & 81.6 & 58.0 & 72.6 \\ \hline
HiT-Tiny \cite{hit}                 & 2023  & 38.4 & 46.7 & 15.3 & 59.6 & 43.4 & 59.6 & 36.5 & 44.2 & 16.9 & 54.6 & 42.3 & 55.8 & 40.3 & 49.3 & 13.8  & 64.7 & 44.4 & 63.2 \\
HiT-Small \cite{hit}                & 2023  & 44.4 & 54.6 & 18.3 & 71.0 & 50.5 & 67.1 & 44.3 & 54.6 & 19.9 & 68.7 & 50.7 & 66.7 & 44.6 & 54.7 & 16.7  & 73.3 & 50.3 & 67.6 \\
HiT-Base \cite{hit}                 & 2023  & 46.6 & 58.8 & 18.7 & 73.4 & 56.9 & 71.8 & 47.3 & 60.1 & 20.5 & 72.4 & 58.3 & 71.7 & 46.0 & 57.5 & 16.8  & 74.4 & 55.5 & 71.8 \\ \hline
LightFC \cite{lightfc}              & 2024  & 43.8 & 53.2 & 15.0 & 72.2 & 55.4 & 66.8 & 44.8 & 55.4 & 16.6 & 71.1 & 56.1 & 68.8 & 42.9 & 51.0 & 13.3  & 73.3 & 54.7 & 64.8 \\
LightFC-vit \cite{lightfc}          & 2024  & 48.7 & 59.7 & 16.2 & 80.8 & 60.0 & 71.3 & 50.7 & 62.9 & 18.2 & 82.3 & 62.9 & 74.3 & 46.7 & 56.5 & 14.2  & 79.3 & 57.1 & 68.3 \\
SMAT \cite{smat}                    & 2024  & 52.3 & 65.8 & 19.2 & 83.1 & 62.4 & 77.7 & 53.3 & 67.5 & 21.5 & 82.6 & 63.3 & \color{blue}\textbf{79.4} & 51.3 & 64.1 & 16.8  & 83.7 & 61.5 & 75.9 \\
LiteTrack-B4 \cite{litetrack}       & 2024  & 52.2 & 67.1 & \color{green}\textbf{24.2} & 79.9 & 62.1 & 76.2 & 53.3 & 68.4 & \color{blue}\textbf{26.0} & 80.8 & 64.5 & 77.7 & 51.0 & 65.8 & 22.5  & 79.1 & 59.7 & 74.6 \\
LiteTrack-B6 \cite{litetrack}       & 2024  & 53.1 & 67.8 & 22.5 & 82.4 & 62.8 & 77.0 & 53.9 & 68.8 & \color{green}\textbf{24.4} & 82.3 & 65.2 & 78.0 & 52.2 & 66.9 & 20.7  & 82.5 & 60.3 & 76.0 \\
LiteTrack-B8 \cite{litetrack}       & 2024  & \color{blue}\textbf{55.0} & \color{blue}\textbf{70.6} & \color{blue}\textbf{24.4} & \color{blue}\textbf{84.6} & \color{green}\textbf{63.8} & \color{blue}\textbf{79.1} & \color{blue}\textbf{55.1} & \color{blue}\textbf{71.0} & \color{blue}\textbf{26.0} & \color{green}\textbf{83.5} & \color{green}\textbf{65.7} & \color{green}\textbf{79.1} & \color{blue}\textbf{54.8} & \color{blue}\textbf{70.2} & \color{green}\textbf{22.8}  & \color{blue}\textbf{85.6} & 61.9 & \color{blue}\textbf{79.0} \\
LiteTrack-B9 \cite{litetrack}       & 2024  & \color{green}\textbf{54.3} & \color{green}\textbf{70.0} & 23.0 & \color{green}\textbf{84.2} & \color{blue}\textbf{65.7} & \color{green}\textbf{78.5} & \color{green}\textbf{54.7} & \color{green}\textbf{70.3} & 23.9 & \color{blue}\textbf{84.3} & \color{blue}\textbf{68.1} & 78.4 & \color{green}\textbf{53.9} & \color{green}\textbf{69.6} & 22.2  & \color{green}\textbf{84.1} & \color{blue}\textbf{63.2} & \color{green}\textbf{78.6} \\
MCITrack-T224 \cite{mcitrack}       & 2024  & 48.4 & 61.8 & 23.4 & 73.6 & 57.8 & 72.3 & 47.2 & 60.7 & 23.2 & 70.7 & 56.7 & 70.2 & 49.6 & 62.9 & \color{red}\textbf{23.6}  & 76.4 & 58.9 & 74.4 \\
MCITrack-S224 \cite{mcitrack}       & 2024  & 49.7 & 62.6 & 23.4 & 76.4 & 60.6 & 73.4 & 47.3 & 59.9 & 23.3 & 71.3 & 58.0 & 69.4 & 52.1 & 65.3 & \color{blue}\textbf{23.5}  & 81.6 & \color{green}\textbf{63.1} & 77.1 \\ \hline
STFTrack-S256                       & -     & \color{red}\textbf{57.6} & \color{red}\textbf{73.8} & \color{red}\textbf{24.5} & \color{red}\textbf{89.9} & \color{red}\textbf{68.2} & \color{red}\textbf{81.2} & \color{red}\textbf{58.9} & \color{red}\textbf{75.1} & \color{red}\textbf{27.0} & \color{red}\textbf{89.7} & \color{red}\textbf{70.7} & \color{red}\textbf{82.1} & \color{red}\textbf{56.3} & \color{red}\textbf{72.5} & 22.1  & \color{red}\textbf{90.1} & \color{red}\textbf{65.7} & \color{red}\textbf{80.2} \\ \hline
\end{tabular}
\end{table*}

%% file: tables/ablation_dataset.tex
\begin{table}[]
\caption{Ablation of used training dataset. The FEM, MTFM, and OTCM modules are disabled in the model.}
\label{table ablation datasets}
\centering
\renewcommand{\arraystretch}{1.1}
\setlength{\tabcolsep}{2pt}
\begin{tabular}{cl|ccc|ccc}
\hline
\multicolumn{2}{c|}{\multirow{2}{*}{}}                                                      & \multicolumn{3}{c|}{SonarT-Fan} & \multicolumn{3}{c}{SonarT-Square} \\ \cline{3-8} 
\multicolumn{2}{c|}{}                                                                       & SR       & PR        & NPR      & SR        & PR         & NPR       \\ \hline
\multicolumn{2}{c|}{Baseline \cite{litetrack}}                                              & 55.1     & 83.5      & 65.7     & 54.8      & 85.6       & 61.9     \\ \hline
\multicolumn{1}{c|}{\multirow{3}{*}{Positive}} & + LaSOT \cite{lasot}                       & 53.4     & 82.9      & 64.1     & 53.7      & 83.7       & 61.0     \\
\multicolumn{1}{c|}{}                          & + GOT10K \cite{got10k}                     & 55.3     & 84.0      & 65.5     & 54.1      & 84.4      &  62.1    \\
\multicolumn{1}{c|}{}                          & + UATD \cite{uatd}   & \textbf{55.9}& \textbf{85.1} & \textbf{66.8} & \textbf{54.9} & \textbf{85.4} & \textbf{63.9} \\ \hline
\multicolumn{1}{c|}{\multirow{3}{*}{Negative}} & Positive + COCO \cite{coco}                & 55.2     & 83.8      & 66.3     & 54.0      & 85.0       & 62.1     \\
\multicolumn{1}{c|}{}                          & Positive + TrackingNet \cite{trackingnet}  & 55.3     & 84.7      & 66.5     & 54.2      & 84.6       & 62.4     \\
\multicolumn{1}{c|}{}                          & Positive + SARDet \cite{sardet}            & 53.9     & 81.7      & 64.2     & 52.0      & 81.6       & 60.1     \\ \hline
\end{tabular}
\end{table}

%% file: tables/ablation_freq_enhance.tex
\begin{table}[]
\caption{Ablation of our Frequency Enhancement Module (FEM). To avoid errors caused by incorrect template updates, the MTFM and OTCM modules are disabled in the model.}
\label{table ablation freq enhance}
\centering
\renewcommand{\arraystretch}{1.1}
\setlength{\tabcolsep}{3pt}
\begin{tabular}{cl|ccc|ccc}
\hline
\multicolumn{2}{c|}{\multirow{2}{*}{}}                              & \multicolumn{3}{c|}{SonarT-Fan} & \multicolumn{3}{c}{SonarT-Square} \\ \cline{3-8} 
\multicolumn{2}{c|}{}                                               & SR       & PR        & NPR      & SR        & PR         & NPR       \\ \hline
\multicolumn{2}{c|}{Baseline \cite{litetrack} + Datasets}           & \textbf{55.9}& \textbf{85.1} & \textbf{66.8} & \textbf{54.9} & \textbf{85.4} & \textbf{63.9}    \\ \hline
\multicolumn{1}{c|}{\multirow{3}{*}{Ablation}} & only HighPass      & 56.4     & 85.6      & 67.1     & 55.0      & 85.1       & 64.5     \\
\multicolumn{1}{c|}{}                          & only LowPass       & 55.0     & 83.5      & 64.8     & 54.0      & 84.3       & 61.8     \\ \cline{2-8} 
\multicolumn{1}{c|}{}                          & HighPass + LowPass & \textbf{56.7} & \textbf{86.1} & \textbf{68.3} & \textbf{55.5} & \textbf{86.4} & \textbf{65.1}  \\ \hline
\end{tabular}
\end{table}

%% file: tables/ablation_template_fusion.tex
\begin{table}[]
\caption{Ablation of Multi-view Template Fusion module (MTFM).}
\label{table ablation template fuision}
\centering
\renewcommand{\arraystretch}{1.1}
\setlength{\tabcolsep}{1.5pt}
\begin{tabular}{cl|ccc|ccc}
\hline
\multicolumn{2}{c|}{\multirow{2}{*}{}}                                                        & \multicolumn{3}{c|}{SonarT-Fan} & \multicolumn{3}{c}{SonarT-Square} \\ \cline{3-8} 
\multicolumn{2}{c|}{}                                                                         & SR       & PR       & NPR       & SR        & PR        & NPR       \\ \hline
\multicolumn{2}{c|}{Tracking Pipeline}                                                        & \textbf{56.7} & \textbf{86.1} & \textbf{68.3} & \textbf{55.5} & \textbf{86.4} & \textbf{65.1}         \\ \hline
\multicolumn{1}{c|}{\multirow{2}{*}{\thead{Dual \\templates}}}        & + Cross Attention     & 57.3     & 87.4      & 69.4     & 56.0      & 87.1        & 65.8     \\
\multicolumn{1}{c|}{}                                                 & + Linear              & 57.9     & 87.6      & 70.5     & 56.2      & 87.4        & 65.7     \\ \hline
\multicolumn{1}{c|}{\multirow{3}{*}{\thead{Multi-view \\template}}}   & + Cross Attention     & 58.6     & 88.4      & 71.0     & 56.7      & 88.0        & 65.8      \\
\multicolumn{1}{c|}{}                                                 & + Channel Enhancement & 58.1     & 87.7      & 70.2     & 56.1      & 87.3        & 64.9     \\
\multicolumn{1}{c|}{}                                                 & + Spatial Enhancement & \textbf{59.0} & \textbf{88.9} & \textbf{71.9} & \textbf{56.9} & \textbf{88.5} & \textbf{66.1} \\ \hline
\end{tabular}
\end{table}

%% file: tables/ablation_tarj_fusion.tex
\begin{table}[]
\caption{Ablation of optimal trajectory Correction module (OTCM).}
\label{table ablation traj fusion}
\centering
\renewcommand{\arraystretch}{1.1}
\setlength{\tabcolsep}{3pt}
\begin{tabular}{cl|ccc|ccc}
\hline
\multicolumn{2}{c|}{\multirow{2}{*}{}}                                 & \multicolumn{3}{c|}{SonarT-Fan} & \multicolumn{3}{c}{SonarT-Square} \\ \cline{3-8} 
\multicolumn{2}{c|}{}                                                  & SR       & PR        & NPR      & SR        & PR         & NPR       \\ \hline
\multicolumn{2}{c|}{Tracking Pipeline + MTFM}                          & \textbf{59.0} & \textbf{88.9} & \textbf{71.9} & \textbf{56.9} & \textbf{88.5} & \textbf{66.1} \\ \hline
\multicolumn{1}{c|}{\multirow{3}{*}{Ablation}} & + Iou Score           & 59.6     & 89.7      & 72.6     & 57.1      & 89.4       & 66.9      \\
\multicolumn{1}{c|}{}                          & + Brightness Response & 59.9     & 90.2      & 73.0     & 57.4      & 90.0       & 67.5      \\
\multicolumn{1}{c|}{}                          & + IoB score           & \textbf{60.0} & \textbf{90.3} & \textbf{73.1} & \textbf{57.6} & \textbf{90.1} & \textbf{67.5} \\ \hline
\end{tabular}
\end{table}

%% file: tables/ablation_sonar_img.tex
\begin{table}[]
\caption{Ablation of Acoustic Image Enhancement methods. \textit{LOW} represents low-frequency image, \textit{High} represents high-frequency image.}
\label{table ablation sonar img}
\centering
\renewcommand{\arraystretch}{1.1}
\setlength{\tabcolsep}{2pt}
\begin{tabular}{ll|ccc|ccc}
\hline
\multicolumn{2}{c|}{\multirow{2}{*}{}}                                  & \multicolumn{3}{c|}{SonarT-Fan} & \multicolumn{3}{c}{SonarT-Square}   \\ \cline{3-8} 
\multicolumn{2}{c|}{}                                                   & SR       & PR        & NPR      & SR        & PR         & NPR        \\ \hline
\multicolumn{2}{c|}{Tracking Pipeline + MTFM/OTCM}                    & \textbf{60.0} & \textbf{90.3} & \textbf{73.1} & \textbf{57.6} & \textbf{90.1} & \textbf{67.5}  \\ \hline
\multicolumn{1}{l|}{\multirow{2}{*}{Ablation}} & Image + High$\times  1$&  60.3    & 90.9      & 73.6     & 57.7      &  90.2      & 68.3      \\
\multicolumn{1}{l|}{}                          & Image + High$\times  2$& \textbf{60.3} & \textbf{90.9} & \textbf{73.6} & \textbf{58.1} & \textbf{90.8} & \textbf{69.0} \\ \hline
\multicolumn{1}{l|}{\multirow{5}{*}{Variants}} & Image + High$\times  3$& \textbf{60.7}     & \textbf{92.0}      & \textbf{74.0}     & 57.9      & \textbf{91.2}       & 68.0       \\ 
\multicolumn{1}{l|}{}                          & Low                    & 49.2     & 78.9      & 62.3     & 43.4      & 70.8       & 51.0       \\
\multicolumn{1}{l|}{}                          & Low + High$\times  1$  & 59.1     & 91.0      & 74.6     & 55.8      & 88.7       & 67.4       \\
\multicolumn{1}{l|}{}                          & Low + High$\times  2$  & 59.1     & 91.5      & 74.7     & 55.9      & 89.3       & 67.9       \\ \cline{2-8} 
\multicolumn{1}{l|}{}                          & Laplacian Sharpening   & 57.9     & 89.3      & 70.7     & 54.8      & 86.7       & 65.6       \\ \hline
\end{tabular}
\end{table}

%% file: figs/vis_tracking_results.tex
\begin{figure*}
	\centering
        \includegraphics[width=18cm]{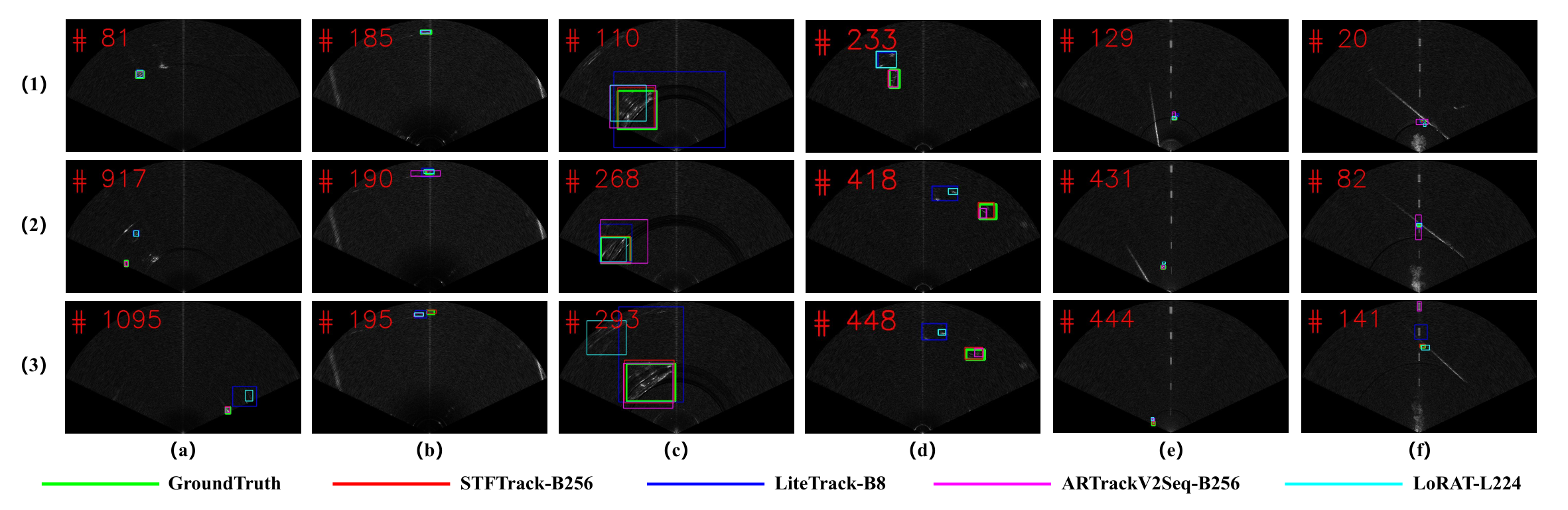}
	\caption{Visualized comparisons of STFTrack-B with LiteTrack \cite{litetrack}, ARTrackV2-Seq \cite{artrackv2}, and LoRAT-L224 \cite{lorat} on six sequenses from SonarT165 dataset. (a) SonarT\_sequence\_001\_fan. (b) SonarT\_sequence\_030\_fan. (c) SonarT\_sequence\_060\_fan. (d) SonarT\_sequence\_100\_fan. (e) SonarT\_sequence\_130\_fan. (f) SonarT\_sequence\_140\_fan.}
	\label{fig: trackingresults}
\end{figure*}

%% file: figs/vis_heatmap.tex
\begin{figure}[]
	\centering
    \includegraphics[width=9cm]{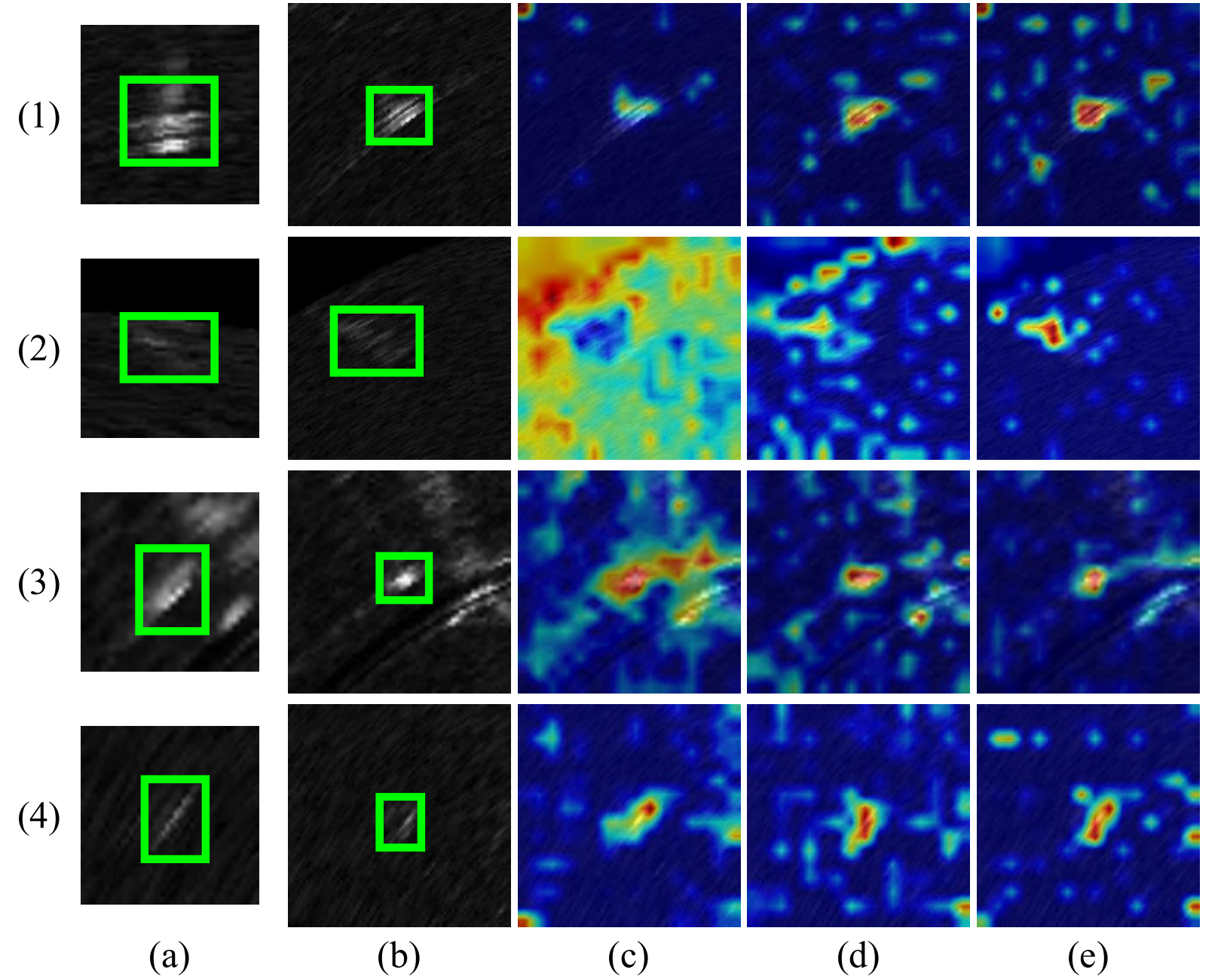}
	\caption{Visualization of heat maps of our STFTrack-B, STFTrack-S, and LiteTrack-B8 \cite{litetrack}. (a) Template. (b) Search Area. (c) LiteTrack-B8 \cite{litetrack}. (d) STFTrack-S. (e) STFTrack-B. (1) SonarT\_sequence\_001\_fan (\# 99). (2) SonarT\_sequence\_020\_fan (\# 51). (3) SonarT\_sequence\_080\_fan (\# 15). (4) SonarT\_sequence\_160\_fan (\# 88). }
	\label{fig: heatmap}
\end{figure}

%% file: figs/vis_failure_cases.tex
\begin{figure*}
	\centering
        \includegraphics[width=18cm]{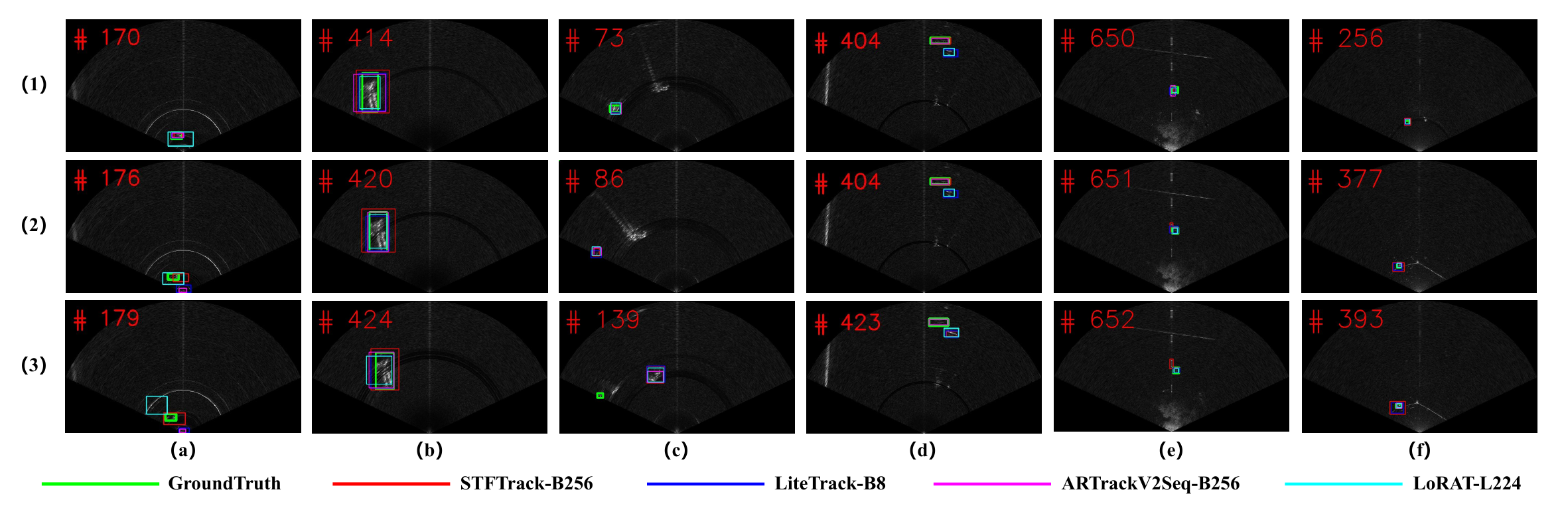}
	\caption{Visualizion of failure cases of STFTrack-B with LiteTrack \cite{litetrack}, ARTrackV2-Seq \cite{artrackv2}, and LoRAT-L224 \cite{lorat} on six sequenses from SonarT165 dataset. (a) SonarT\_sequence\_015\_fan. (b) SonarT\_sequence\_055\_fan. (c) SonarT\_sequence\_065\_fan. (d) SonarT\_sequence\_090\_fan. (e) SonarT\_sequence\_125\_fan. (f) SonarT\_sequence\_160\_fan.}
	\label{fig: fail cases}
\end{figure*}

%% file: sections/6_discussion.tex
\section{Discussion}
\input{tables/discussion_speed}

\subsection{Application Potential}
We further explore the potential applications of the proposed method. As shown in Table \ref{table discussion speed}, we report the performance, parameters, FLOPs, and speed of STFTrack and its baseline LiteTrack \cite{litetrack}. Compared to the baseline LiteTrack \cite{litetrack}, our STFTrack has only a slight speed penalty in model inference, which means that our method has great potential for application. In addition, although we introduce template update and trajectory post-processing, the former usually only calculates once using the MTFM module when the template is updated, while the latter only performs batch-based fast calculation when the trajectory is abnormal. Therefore, the introduction of these modules will not affect the speed.

\subsection{Limitation}
Although we introduce a large-scale benchmark dataset for underwater acoustic object tracking, it still contains several shortcomings. First, the proposed benchmark contains only test sequences, making it challenging for current acoustic trackers to learn the discriminative features of acoustic targets. Secondly, the target categories in our benchmark do not fully cover typical underwater targets such as pipeline objects, open-frame remotely operated vehicles (ROVs), etc. Third, our benchmark needs more field environment sequences, such as acoustic environments in lakes and acoustic environments in the ocean.

In the future, we will prepare a more comprehensive range of underwater object types and conduct ocean experiments to collect more diverse and scene-rich underwater acoustic object tracking datasets.

\subsection{Expansion of Acoustic Vision}
Acoustic images utilize the acoustic reflection characteristics of targets to form visual images. In underwater environments, forward-looking sonar is commonly used to construct acoustic images of targets, also known as sonar images. However, there are also acoustic vision tasks in other fields, such as medical image processing, where the use of ultrasound to detect human tissue also requires processing of acoustic images. Therefore, exploring acoustic (sonar) image processing methods also has reference significance for other acoustic tasks (For example, \cite{tipsm} explores a B-mode ultrasound tracker for medical image processing).

%% file: tables/discussion_speed.tex
\begin{table}[]
\caption{The performance, parameters, GFLOPs, and speed of STFTrack and the baseline method. Here we only count the statistics during tracking inference. The GPU is NVIDIA RTX 3090Ti. The OrinNX is NVIDIA Orin NX.}
\label{table discussion speed}
\centering
\renewcommand{\arraystretch}{1.1}
\setlength{\tabcolsep}{3pt}
\begin{tabular}{l|c|c|c|cc}
\hline
\multirow{2}{*}{} & Performance & \multirow{2}{*}{Params} & \multirow{2}{*}{FLOPs} & \multicolumn{2}{c}{Speed} \\ \cline{2-2} \cline{5-6} 
                                & SR  /  PR       &                 &               & GPU        & OrinNX       \\ \hline
STFTrack-S                      & 57.6 / 89.9     & 46.1M           & 10.1G         & 283        & 25           \\
STFTrack-B                      & 59.2 / 90.8     & 56.7M           & 12.8G         & 222        & 21           \\ \hline
LiteTrack-B6 \cite{litetrack}   & 53.1 / 84.6     & 39.0M           & 10.1G         & 288        & 26          \\
LiteTrack-B8 \cite{litetrack}   & 55.0 / 82.4     & 49.6M           & 12.8G         & 226        & 22           \\ \hline
\end{tabular}
\end{table}

%% file: sections/7_conclusion.tex
\section{Conclusion}
In this work, we propose a large-scale underwater acoustic object tracking (UAOT) benchmark SonarT165. SonarT165 contains 165 square sequences and 165 fan sequences, totaling 205K annotations. It reflects the characteristics of acoustic images and the typical challenges of sonar object tracking.
We evaluate popular general trackers and lightweight trackers on the benchmark, and experimental results show that SonarT165 poses a challenge to these trackers. In addition, we propose STFTrack-B and STFTrack-S to deal with the issues of target appearance changes and interference in UAOT. STFTrack introduces a multi-view template fusion module and an optimal trajectory correction module. The former achieves multi-view dynamic template modeling and spatio-temporal target appearance modeling. The latter achieves the correction of suboptimal matching between kalman filter predicted boxes and candidate bounding boxes. Extensive experiments show that STFTrack achieves state-of-the-art performance. Comprehensive experiments show that STFTrack achieves state-of-the-art performance.